\documentclass[11pt, preprint]{imsart}
\RequirePackage[OT1]{fontenc}
\RequirePackage{amsthm,amsmath}
\RequirePackage[colorlinks,citecolor=blue,urlcolor=blue]{hyperref}
\usepackage{graphicx}
\usepackage{enumerate}
\usepackage{latexsym}
\usepackage{subfigure}
\usepackage{bm}
\usepackage{amssymb}
\usepackage{multirow}
\usepackage{float}
\usepackage{paralist}
\usepackage{natbib}
\usepackage{mathtools}
\usepackage{tikz}
\usetikzlibrary{automata,arrows,calc,positioning}

\usepackage[inline]{enumitem}
\usepackage{threeparttable}
{\end{quotation}}
\usepackage[normalem]{ulem}

\numberwithin{equation}{section}

\def \mcF{\mathcal{F}}
\def \mcH{\mathcal{H}}

\DeclareMathOperator*{\argmin}{arg\,min}

\numberwithin{equation}{section}

\def \mcF{\mathcal{F}}
\def \mcH{\mathcal{H}}

\def \mcNN{\mathcal{N}\mathcal{N}}

\def \mcNN{\mathcal{N}\mathcal{N}}
\def \bfx{\mathbf{x}}
\def \bfu{\mathbf{u}}
\def \bfD{\mathbf{D}}
\def \bfX{\mathbf{X}}
\def \bfB{\mathbf{B}}
\def \bfi{\mathbf{i}}
\def \bfY{\mathbf{Y}}
\def \bfy{\mathbf{y}}
\def \bff{\mathbf{f}}
\def \bfv{\mathbf{v}}
\def \bfp{\mathbf{p}}
\def \bfE{\mathbf{E}}
\def \bfP{\mathbf{P}}
\def \bfepsilon{\bm{\epsilon}}
\def \ev{\mathbb{E}}
\def \pr{\mathbb{P}}
\def \mcC{\mathcal{C}}

\newcommand{\floor}[1]{\left\lfloor #1 \right\rfloor}

\usepackage{pifont}
\newcommand{\xmark}{\textrm{\ding{53}}}%

\newcommand{\vo}{\vec{o}\@ifnextchar{^}{\,}{}}
\startlocaldefs
\theoremstyle{plain}

\newtheorem*{remark}{Remark}
\newtheorem{theorem}{\indent Theorem}
\newtheorem*{theorem*}{\indent Theorem}

\newtheorem{Assumption}{Assumption}

\newtheorem{lemma}{\indent Lemma}
\newtheorem{proposition}{\indent Proposition}

\theoremstyle{definition}

\makeatletter\def\theequation{\arabic{section}.\arabic{equation}}
\@addtoreset{equation}{section}\makeatother

\endlocaldefs

\begin{document}
\begin{frontmatter}
\maketitle
\title{Optimal Nonparametric Inference via Deep Neural Network}

\runtitle{Optimal Nonparametric Inference via Deep Neural Network}

	\begin{aug}
		\author{\fnms{Ruiqi} \snm{Liu}\thanksref{e1}\ead[label=e4]{liuruiq@iu.edu}},
		\author{\fnms{Ben} \snm{Boukai}\thanksref{m1}\ead[label=e1]{bboukai@iupui.edu}}
		\and
		\thankstext{e1}{Department of Mathematics and Statistics, Texas Tech University, USA.}
		\thankstext{m1}{Department of Mathematical Sciences, Indiana University–Purdue University Indianapolis, USA.}
		\author{\fnms{Zuofeng} \snm{Shang}\thanksref{t1}
			\ead[label=e3]{shangzf@iu.edu}}
         \thankstext{t1}{Corresponding author. Department of Mathematical Sciences, New Jersey Institute of Technology, USA.
         Email: zshang@njit.edu.}
         \runauthor{Liu et al.}

	\end{aug}

 \begin{abstract}
Deep neural network is a state-of-art method in modern science and technology. 
Much statistical literature have been devoted to understanding its 
performance in nonparametric estimation, whereas the results are suboptimal due to a redundant logarithmic sacrifice.
In this paper, we show that such log-factors are not necessary.
We derive upper bounds for the $L^2$ minimax risk in nonparametric estimation. 
Sufficient conditions on network architectures are provided such that the upper bounds become optimal
(without log-sacrifice). 
Our proof relies on an explicitly constructed network estimator based on tensor product B-splines.
We also derive asymptotic distributions for the constructed network and a relating hypothesis testing procedure.
The testing procedure is further proved as minimax optimal under suitable network architectures.
\end{abstract}

\end{frontmatter}

\section{Introduction}

With the remarkable development of modern technology, difficult learning problems can nowadays be tackled smartly via deep learning architectures.
For instance, deep neural networks have led to impressive performance  
in fields such as computer vision, natural language processing, image/speech/audio recognition, social network filtering, machine translation, bioinformatics, drug design, medical image analysis,
where they have demonstrated superior performance to human experts. 
The success of deep networks hinges on their rich expressiveness
(see \cite{db11} , \cite{rpkgdj17}, \cite{mpcb14}, \cite{bs14}, \cite{t16}, \cite{l17} and \cite{y17, y18}).
Recently, deep networks have played an increasingly important role in statistics
particularly in nonparametric curve fitting (see \cite{kk05, hk06, kk17, km11, s19}).
Applications of deep networks in other fields such as image processing or pattern recgnition include, to name a few, \cite{lbh15}, \cite{dlhyysszhw13}, \cite{wwhwzzl14}, \cite{gg16}, etc.

A fundamental problem in statistical applications of deep networks is how accurate they can estimate a nonparametric regression function. 
To describe the problem, let us consider i.i.d. observations $(Y_i, \bfX_i)$, $i=1,2,\ldots, n$ generated from the following nonparametric model:
\begin{eqnarray}
	Y_i=f_0(\bfX_i)+\epsilon_i, \label{eq:model}
\end{eqnarray}
where $\bfX_i \in [0,1]^d$ are i.i.d. $d$-dimensional predictors for a fixed $d\ge1$, 
$\epsilon_i$ are i.i.d. random noise with $E(\epsilon_i)=0$ and $Var(\epsilon_i)=\tau^2$,
$f_0$ is an unknown function belonging to some function space $\mcH$. 
For any $L\in\mathbb{N}$ and $\bfp=(p_1,\ldots,p_L)\in\mathbb{N}^L$,
let $\mcF(L,\bfp)$ denote the collection of network functions from $\mathbb{R}^d$ to $\mathbb{R}$ consisting
of $L$ hidden layers with the $l$th layer including $p_l$ neurons. 
The problem of interest is to find an order $R_n$ that controls the $L^2$ minimax risk:
\begin{equation}\label{eqn:goal}
	\inf_{\widehat{f}\in\mcF(L,\bfp)}\sup_{f_0\in \mcH}\ev_{f_0}\bigg(\|\widehat{f}-f_0\|_{L^2}^2\bigg|\mathbb{X}\bigg)=O_P(R_n),
\end{equation}
where $\mathbb{X}=\{\bfX_1,\ldots,\bfX_n\}$, the infimum is taken over all estimators 
$\widehat{f}\in\mcF(L,\bfp)$, and $\ev_{f_0}$ represents the expectation taken over the conditional distribution of $Y_i$'s given $\bfX_i$'s
with $Y_i,\bfX_i$ generated from model (\ref{eq:model}), and $O_P$ stands for stochastic boundedness, which will  be formally defined at the end of Section \ref{sec:prelim}. 
In other words, we are interested in the performance of the ``best'' network estimator
in the ``worst'' scenario.

Existing results regarding (\ref{eqn:goal}) are sub-optimal.
For instance, when $\mcH$ is a $\beta$-smooth H\"{o}lder class and $L,\bfp$ are properly selected, 
it has been argued that $R_n=n^{-\frac{2\beta}{2\beta+d}}(\log{n})^s$ for some constant $s>0$;
see \cite{kk05, hk06, kk17, km11, s19, su18, flm18, DIVE2020, FDA_shang2020}. Such results are mostly proved based on empirical processes techniques
in which the logarithmic factors
arise from the entropy bound of the neural network class.
The aim of this paper is to fully remove the redundant logarithmic factors, i.e., 
under proper selections of $L,\bfp$ one actually has $R_n=n^{-\frac{2\beta}{2\beta+d}}$ in (\ref{eqn:goal}).
This means that neural network estimators can exactly achieve minimax estimation rate.
Our proof relies on an explicitly constructed neural network through tensor product B-splines
which is proved minimax optimal.  One technical contribution of this paper is to show that
tensor product B-splines can be effectively expressed by deep networks. Compared with other basis structures such as local Taylor expansions, e.g., \cite{y17} and \cite{s19},
the tensor product B-splines framework is
convenient to our theoretical analysis due to its rich statistical literature; see \cite{h98} and \cite{h03}.

Some interesting byproducts are worth mentioning. First, we will derive the pointwise asymptotic distribution of the constructed neural network estimator 
which will be useful to establish pointwise confidence interval.
Second, the constructed neural network estimator will be further used 
as a test statistic which is proved optimal when $L,\bfp$ are properly selected. 
As far as we know, these are the first provably valid confidence interval and test statistic
based on neural networks in nonparametric regression.
Third, the rate $R_n$ can be further improved when $f_0$ satisfies additional structures.
Specifically, we will show that $R_n=n^{-\frac{2\beta}{2\beta+1}}$ if $f_0$ satisfies
additive structure, i.e., $f_0$ is a sum of univariate $\beta$-H\"{o}lder functions.
Such rate is minimax according to \cite{s85}.

This paper is organized as follows. 
Section \ref{sec:prelim}  includes some preliminaries on
deep networks and defines some notation.
In Section \ref{sec:optimal:rate:convergence}, we derive
upper bounds for the minimax risk and investigate their optimality. 
Section \ref{sec:basis:approximation} provides the proof of the main result,
which covers the construction of (optimal) network and relates results on network approximation
of tensor product B-splines.
As by products, we also provide limiting distribution and optimal testing results in Section \ref{sec:byproducts}. We further study the additive model using network approximation in Section \ref{sec:additive:model}. The Appendix contains the proofs of relevant lemmas and a table indexing some important symbols used in the proof.

\section{Preliminaries and Notation}\label{sec:prelim}
In this section, we review some notion about deep networks and function spaces, as well as provide useful symbols or notation 
used throughout this paper.
Throughout let $\sigma$ denote the rectifier linear unit (ReLU) activation function,
i.e., $\sigma(x)=(x)_+$ for $x\in\mathbb{R}$.
For any real vectors $\bfv=(v_1,\ldots,v_r)^T$ and $\bfy=(y_1,\ldots,y_r)^T$, we define the shift activation function
$\sigma_\bfv(\bfy)=(\sigma(y_1-v_1),\ldots,\sigma(y_r-v_r))^T$.
Let $\bfp=(p_1,\ldots,p_L)\in\mathbb{N}^L$, and we say $f\in\mcF(L,\bfp)$ if 
\[
f(\bfx)=W_{L+1}\sigma_{\bfv_{L}}W_{L}\sigma_{\bfv_{L-1}}\ldots W_2\sigma_{\bfv_1}W_1 \bfx,\,\,\bfx \in \mathbb{R}^{d},
\]
where $\bfv_{l}\in \mathbb{R}^{p_l}$ is a shift vector and $W_{l}\in \mathbb{R}^{p_{l}\times p_{l-1}}$ is a weight matrix, and $\bfx$
represents the argument of $f$. We adopt the representation $\bfx=(x_1,\ldots,x_d)^T$ with $x_j$ being the $j$th component of $\bfx$ and the convention $p_0=d$ and $p_{L+1}=1$.
For simplicity, we only consider fully connected networks and do not make any sparsity assumptions on the entries of $\bfv_l$ and $W_l$.

Next let us review the concept of H\"{o}lder space.
Let $\Omega=[0,1]^d$ denote the domain of the functions. 
For $f$ defined on $\Omega$, we define the supnorm, $L^2$-norm and empirical norm of $f$ by $\|f\|_{\sup}=\sup_{\bfx\in\Omega}|f(\bfx)|$, $\|f\|_{L^2}^2=\int_\Omega f(\bfx)^2Q(\bfx)d\bfx$ and $\|f\|_n^2=\frac{1}{n}\sum_{i=1}^n f(\bfX_i)$, respectively.
Here $Q(\cdot)$ is the probability density for the predictor $\bfX_i$'s.
For any $\bm{\alpha}=(\alpha_1, \alpha_2, \ldots, \alpha_d)\in \mathbb{N}^d$, we denote $|\bm{\alpha}|=\sum_{j=1}^d\alpha_j$ and $\partial^{\bm{\alpha}}f=\frac{\partial^{|\bm{\alpha}|}f}{\partial x_1^{\alpha_1}\ldots \partial x_d^{\alpha_d}}$ whenever the partial derivative exists.  For any $\beta>0$ and $F>0$, let $\Lambda^\beta(F, \Omega)$ denote the ball of $\beta$-H\"{o}lder functions with radius $F$, i.e.,
\begin{eqnarray}
	\Lambda^\beta(F, \Omega)=\bigg\{f: \Omega\to \mathbb{R} \bigg|\;\sum_{\bm{\alpha}:|\bm{\alpha}|\leq \floor{\beta}}\|\partial^{\bm{\alpha}}f\|_{\sup}+\sum_{\bm{\alpha}:|\bm{\alpha}|=\floor{\beta}}\sup_{\bfx_1\neq \bfx_2\in \Omega}\frac{|\partial^{\bm{\alpha}}f(\bfx_1)-\partial^{\bm{\alpha}}f(\bfx_2)|}{\|\bfx_1-\bfx_2\|_2^{\beta-\floor{\beta}}}\leq F\bigg\},\nonumber
\end{eqnarray}
in which $\floor{\beta}$ is the largest integer strictly smaller than $\beta$.

At the end, we need some notation for vector, matrix and asymptotic analysis. For vector  $v=(v_1, \ldots, v_p)\in \mathbb{R}^p$, let $\|v\|_\infty=\max_{1\leq i \leq p}|v_i|$ and $\|v\|_2=\sqrt{\sum_{i=1}^p v_i^2}$ be its supnorm and Euclidean norm.  Let $\lambda_{\min}(\cdot)$ and $\lambda_{\max}(\cdot)$  denote the smallest and largest eigenvalues of a squared matrix.  For two positive sequences $a_n$ and $b_n$, we define $a_n\asymp b_n$ if there exist positive constants $C_1,C_2$
such that $C_1a_n\le b_n\le C_2a_n$. We say a sequence of random variables  $X_n=O_P(a_n)$ for some positive deterministic sequence $a_n$  if for any $\varepsilon>0$, there exists a constant $C_\varepsilon>0$ such that $P(|X_n|\ge C_\varepsilon a_n)\le \varepsilon$ 
for all $n\ge1$. Finally, we denote $X_n=o_P(a_n)$ if $\lim_{n\to \infty}P(|X_n|>\epsilon a_n)=0$ for any $\epsilon>0$. 

\section{Minimax Neural Network Estimation}\label{sec:optimal:rate:convergence}

In this section, we derive an upper bound for the $L^2$ minimax risk in the problem (\ref{eqn:goal}).
The risk bound will be proved optimal under suitable circumstances.
To simplify the expressions, we only consider networks with architecture $(L,\bfp(T))$,
where $\bfp(T):=(T,\ldots,T)\in\mathbb{N}^L$ for any $T\in\mathbb{N}$.
In other words, we focus on networks with $L$ hidden layers and each having $T$ neurons. 
Our results hold under suitable conditions on $L$ and $T$ as well as the following assumption on the design and model error.
\begin{Assumption}\label{A0}
The probability density $Q(\bfx)$ of $\mathbf{X}$ is supported on $\Omega$.
There exists a constant $c>0$ such that $c^{-1}\le Q(\bfx)\le c$ for any $\bfx\in\Omega$.
The error terms $\epsilon_i$'s are independent of $\bfX_i$'s. 
\end{Assumption}

\begin{theorem}\label{thm:neural:estimator:rate:of:convergence:main:text}
Let Assumption \ref{A0} be satisfied. Suppose that $L\to \infty$, $T\to\infty$ and $T\log T=o(n)$ as $n\to\infty$, then for any fixed constants $\beta, F>0$,
it follows that
\begin{eqnarray}\label{thm1:risk:bound}
	\inf_{\widehat{f}\in \mathcal{F}(L,\bfp(T))}\sup_{f_0\in \Lambda^\beta(F, \Omega)}
	\ev_{f_0}\bigg(\|\widehat{f}-f_0\|_{L^2}^2\bigg|\mathbb{X}\bigg)=O_P\bigg(\frac{1}{T^{\frac{2\beta}{d}}}+\frac{T}{n}+\frac{T^2}{4^{\frac{L}{d+k}}}\bigg),
\end{eqnarray}
where $k$ is the smallest integer satisfying $k\ge \max(\beta, 2)$. As a consequence, if $T\asymp n^{\frac{d}{2\beta+d}}$ and $n^{\frac{2\beta+2d}{2\beta+d}}=O(4^{\frac{L}{d+k}})$, then the following holds:
\begin{eqnarray*}
	\inf_{\widehat{f}\in \mathcal{F}(L,\bfp(T))}\sup_{f_0\in \Lambda^\beta(F, \Omega)}
	\ev_{f_0}\bigg(\|\widehat{f}-f_0\|_{L^2}^2\bigg|\mathbb{X}\bigg)=O_P(n^{-\frac{2\beta}{2\beta+d}}).
\end{eqnarray*}
\end{theorem}
The $O_P$ in Theorem \ref{thm:neural:estimator:rate:of:convergence:main:text} represents stochastic boundedness as defined at the end of Section \ref{sec:prelim}, which involves some fixed constant.  The constant term in $O_P$
relies on $c$ (Assumption \ref{A0}), $\beta$ (smoothness of $f_0$), $k$ (auxiliary integer related to $\beta$), $F$ (radius of function space), 
and $d$ (dimension of the design point), and is free of $n,T,L$. We ignore the constant term as the focus of this paper is to
investigate the impact of $T,L$ (network architecture) and $n$ (sample size) on the minimax rate. Moreover, the choice $T\asymp n^{\frac{d}{2\beta+d}}$ would be satisfied if $T=c_0n^{\frac{d}{2\beta+d}}$ for some fixed constant $c_0>0$.
 
Proof of Theorem \ref{thm:neural:estimator:rate:of:convergence:main:text} 
relies on an explicitly constructed network estimator
based on tensor product B-splines of order $k$, where $k\geq \max(\beta, 2)$ is the constant specified in condition of Theorem \ref{thm:neural:estimator:rate:of:convergence:main:text}.
The minimax risk bound in (\ref{thm1:risk:bound}) consists of three components ${T^{-\frac{2\beta}{d}}}, n^{-1}T, 4^{-\frac{L}{d+k}}T^2$
corresponding to the bias, variance and approximation error of the constructed network.
The optimal risk bound is achieved through balancing the three terms.
The approximation error of the constructed network decreases exponentially along with $L$. 
Networks constructed based on other methods
such as local Taylor approximations (\cite{y17}, \cite{y18} and \cite{s19} have similar approximation performance.
However, their statistical properties are more challenging to deal with due to the unbalanced eigenvalues of the corresponding basis matrix.
In contrast, the eigenvalues of the tensor product B-spline basis matrix are known to have balanced orders, e.g., see  \cite{d78},
which plays an important role in deriving the risk bound.
Also notice that the risk bound will blow out when $L$ is fixed,
which partially explains the superior performance of deep networks compared with shallow ones; see \cite{es16}.

\section{Construction of Optimal Networks}\label{sec:basis:approximation}
In this section, we prove Theorem \ref{thm:neural:estimator:rate:of:convergence:main:text} by explicitly constructing a network estimator $\widehat{f}_{\textrm{net}}\in\mathcal{F}(L,\bfp(T))$ and deriving its risk bound.
The construction process starts from a pilot estimator $\widehat{f}_{\textrm{pilot}}$ obtained under 
tensor product B-splines with order $k\geq \max(\beta, 2)$.
The tensor product B-spline basis functions are further approximated through explicitly constructed multi-layer networks,
which will be aggregated to obtain the network estimator $\widehat{f}_{\textrm{net}}$.
The key step is to show that the discrepancies between the tensor product B-spline basis functions and the corresponding network approximations are reasonably small
such that $\widehat{f}_{\textrm{net}}$ will perform similarly as $\widehat{f}_{\textrm{pilot}}$, and thus, optimally. 

Our construction is different from \cite{y17} and \cite{s19}, where the basis functions are obtained through local Taylor approximation. 
We find that the eigenvalue performance of the local Taylor basis matrix is difficult to quantify so that the corresponding pilot estimator 
cannot be used effectively. Instead, the pilot estimator based on tensor product B-splines is more convenient to deal with.
Other basis such as wavelets or smoothing splines may also work but this will be explored elsewhere.
\subsection{A Pilot Estimator Through Tensor Product B-splines}\label{section:pilot:bspline}
In this subsection, we review tensor product $B$-splines and construct the corresponding pilot estimator.
For any integer $M\geq 2$, 
let $0=t_0<t_1<\cdots<t_{M-1}<t_M=1$ be knots that form a partition of the unit interval.
The definition of univariate B-splines of order $k\ge2$ depends on 
additional knots $t_{-k+1}<t_{-k+2}<\ldots<t_{-1}<0$ and $1<t_{M+1} <\ldots< t_{M+k-1}$. Given knots $t=(t_{-k+1},\ldots, t_{M+k-1})\in \mathbb{R}^{M+2k-1}$, the univariate $B$-spline basis functions of order $k$, denoted $B_{i,k}(x)$, $i=-k+1,-k+2,\ldots, M-1$, can be defined inductively by $B_{i,s}(x)$ for $s=2,3,\ldots, k$. For $s=2$ and $-k+1\le i\le M+k-3$, define
\begin{eqnarray*}
	B_{i, 2}(x)=\begin{cases}
		\frac{x-t_i}{t_{i+1}-t_i}, & \textrm{if } x\in [t_i, t_{i+1}]\\
		\frac{t_{i+2}-x}{t_{i+2}-t_{i+1}},  & \textrm{if } x\in [t_{i+1}, t_{i+2}]\\
		0,& \textrm{elsewhere}
	\end{cases}.
\end{eqnarray*}
Suppose that $B_{i,s}(x)$, $i=-k+1,\ldots, M+k-s-1$ have been defined. We recursively define
\begin{eqnarray}\label{induction:formula}
	B_{i, s+1}=a_{i,s}B_{i,s,t}+b_{i,s}B_{i+1,s, t},\,\,\textrm{for $i=-k+1,-k+2,\ldots, M+k-s-2$,}
\end{eqnarray}
where 
\begin{eqnarray*}
	{a}_{i,s}(x)=\begin{cases}
	0, & \textrm{if } x<t_i\\
	\frac{x-t_i}{t_{i+s}-t_i}, &\textrm{if } t_i\leq x \leq t_{i+s}\\
	0,&\textrm{if } x> t_{i+s}\\
	\end{cases},\;\;\;{b}_{i,s}(x)=\begin{cases}
	0, & \textrm{if } x<t_{t+1}\\
	\frac{t_{i+s+1}-x}{t_{i+s+1}-t_{i+1}}, &\textrm{if } t_{i+1}\leq x \leq t_{i+s+1}\\
	0,&\textrm{if } x> t_{i+s+1}\\
	\end{cases}.
\end{eqnarray*}
Proceeding with this construction, we can obtain $B_{i,k}(x)$.

To approximate a multivariate function, we adopt the tensor product $B$-splines.
Let $\Gamma=\{-k+1, -k+2, \ldots, 0, 1, \ldots, M-1\}^d$ and $q=|\Gamma|=(M+k-1)^d$. 
For $\bfi=(i_1, i_2, \ldots, i_d)\in \Gamma$, we define 
$D_{\bfi,k}(\bfx)=\prod_{j=1}^dB_{i_j,k}(x_j)$
and obtain the corresponding pilot estimator 
\begin{eqnarray}
	\widehat{f}_{\textrm{pilot}}(\bfx)=\sum_{\bfi \in \Gamma}\widehat{b}_{\bfi}D_{\bfi, k}(\bfx),\label{eq:pilot:estimator}
\end{eqnarray}
where $\{\widehat{b}_{\bfi}, \bfi\in \Gamma\}$ are the basis coefficients obtained 
by the following least square estimation:
\begin{equation}\label{LSE:eqn}
\widehat{C}:=[\widehat{b}_{\bfi}]_{\bfi\in\Gamma}=\argmin_{b_\bfi, \bfi \in \Gamma}\sum_{i=1}^n \left(Y_i-\sum_{\bfi\in\Gamma}b_\bfi D_{\bfi,k}(\bfX_i)\right)^2.
\end{equation}

\subsection{Network Approximation of Tensor Product B-splines}
In this subsection, we approximate $D_{\bfi,k}$'s through multilayer neural networks.
We first construct networks that approximate the univariate B-spline basis $B_{i,k}$'s,
and then multiply these networks through a product network $\xmark_s$ introduced by \cite{y17} to approximate the tensor product B-spline basis. Here, the product network
$\xmark_s(x_1, x_2, \ldots, x_s)$ is constructed to approximate the monomials $\prod_{j=1}^s x_j$. Unlike \cite{y17} and \cite{s19}, our construction proceeds in an inductive manner due to the intrinsic induction structure of B-splines.

To proceed, let us introduce some notation. For $L,p_0,\ldots,p_{L+1}\in\mathbb{N}$, let us denote $\mcNN(L,(p_0,p_1,$ $\ldots,p_L,p_{L+1}))$
as the class of $p_0$-input-$p_{L+1}$-output ReLU neural network functions of $L$ hidden layers,
with the $j$th layer consisting of $p_j$ nodes, for $j=1,\ldots,L$. In particular, with $p_0=d$ and $p_{L+1}=1$, $\mcNN(L, (p_0,p_1,\ldots,p_L,p_{L+1}))$ is equivalent to $\mcF(L, (p_1,\ldots, p_{L+1}))$. 
The following Propositions \ref{proposition:appriximation:square}-\ref{proposition:appriximation:product:k} quantify the approximation error of the product network $\xmark_s$.
\begin{proposition}\label{proposition:appriximation:square}
For any integer $m\geq 1$, there exists $SQ \in \mcNN(2m, (1,4,\ldots,4,1))$ such that
\begin{eqnarray*}
|SQ(x)-x^2|\leq 2^{-2m-2}, \;\;\; \textrm{ for all } x\in [0, 1].
\end{eqnarray*}
\end{proposition}
\begin{proof}[{\it Proof of Proposition \ref{proposition:appriximation:square}}]
For $s\ge1$, let $g, g_s$ be functions taking values in $[0, 1]$ defined as
\begin{eqnarray*}
	g(x)=\begin{cases}
	2x, & \textrm{if } 0\leq x<1/2\\
	2(1-x), & \textrm{ if } 1/2 \leq x\geq \leq 1
	\end{cases}, \;\;\;\;	  g_s=\underbrace{g\circ g\circ \cdots g}_\textrm{$s$ times}.
\end{eqnarray*}
It can be shown by induction that
\begin{eqnarray*}
	g_s(x)=\begin{cases}
	2^s\left(x-\frac{2k}{s^2}\right),& \textrm{if } x\in [\frac{2k}{2s}, \frac{2k+1}{2^s}]\\
	2^s\left(\frac{2k}{2^s}-x\right),& \textrm{if } x\in [\frac{2k-1}{2s}, \frac{2k}{2^s}]
	\end{cases}.
\end{eqnarray*}
Let $h_m(x)$ be the linear interpolation of $h(x)=x^2$ at points $k2^{-m}$, for $k=0,1,\ldots, 2^m$.
Namely,
\begin{eqnarray*}
h_m(x)=\frac{2k+1}{2^m}x-\frac{k(k+1)}{4^m}, \quad \textrm{if } x \in [k2^{-m}, (k+1)2^{-m}].
\end{eqnarray*}
By direct examinations, we have
\begin{eqnarray*}
	|h(x)-h_m(x)|\leq 2^{-2m-2}, \quad \textrm{for all } x\in [0,1].
\end{eqnarray*}
Moreover, by induction, it can be shown that
\begin{eqnarray*}
	h_{m-1}(x)-h_m(x)=\frac{g_m(x)}{4^m}, \quad \textrm{for all } x\in [0,1].
\end{eqnarray*}
The above equation and the fact that $h_0(x)= x$ lead to 
\begin{eqnarray*}
	h_m(x)=x-\sum_{s=1}^m\frac{g_s(x)}{4^s}.
\end{eqnarray*}
Since $g(x)=2\sigma(x)-4\sigma(x-\frac{1}{2})+2\sigma(x-1)$,
$g(x)$ is a neural network consisting of one hidden layer. 
Define $SQ=h_m$, then $SQ$ is a single-input-single-output neural network of $2m$ hidden layers, and 
each layer contains 4 neurons, i.e., $SQ\in \mcNN(2m, (1,4,\ldots, 4,1))$; see Figure \ref{figure:construction:SQ:m3} for the case when $m=3$.
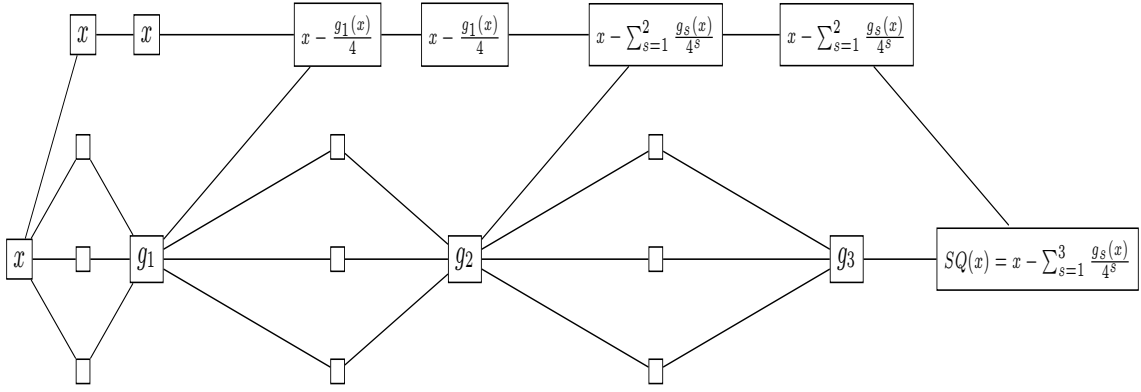
\begin{figure}[ht!]
\centering
 \resizebox {6 in} {2 in} {
\begin{tikzpicture}[>=stealth', shorten >=1pt, auto,
    node distance=2.5cm, scale=1, 
    transform shape, align=center, 
    state/.style={circle, draw, minimum size=0.5cm}, small/.style={draw}]
\node[small] (x) at (1,2) {$x$};
\node[small]  (box11) at (2,1) {};
\node[small]  (box12) at (2,2) {};
\node[small]  (box13) at (2,3) {};
\node[small]  (box14) at (2,4) {$x$};

\node[small]  (box21) at (3,2) {$g_1$};
\node[small]  (box22) at (3,4) {$x$};

\node[small]  (box31) at (6,1) {};
\node[small]  (box32) at (6,2) {};
\node[small]  (box33) at (6,3) {};
\node[draw, font=\tiny]  (box34) at (6,4) {$x-\frac{g_1(x)}{4}$};

\node[small]  (box41) at (8,2) {$g_2$};
\node[draw, font=\tiny]  (box42) at (8,4) {$x-\frac{g_1(x)}{4}$};

\node[small]  (box51) at (11,1) {};
\node[small]  (box52) at (11,2) {};
\node[small]  (box53) at (11,3) {};
\node[draw, font=\tiny]  (box54) at (11,4) {$x-\sum_{s=1}^2\frac{g_s(x)}{4^s}$};

\node[small]  (box61) at (14,2) {$g_3$};
\node[draw, font=\tiny]  (box62) at (14,4) {$x-\sum_{s=1}^2\frac{g_s(x)}{4^s}$};

\node[draw, font=\tiny] (box71) at (17,2) {$SQ(x)=x-\sum_{s=1}^3\frac{g_s(x)}{4^s}$};
\draw (x)--(box11)
(x)--(box12)
(x)--(box13)
(x)--(box14)
(box11)--(box21)
(box12)--(box21)
(box13)--(box21)
(box14)--(box22)

(box21)--(box31)
(box21)--(box32)
(box21)--(box33)
(box21)--(box34)
(box22)--(box34)

(box31)--(box41)
(box32)--(box41)
(box33)--(box41)
(box34)--(box42)

(box41)--(box51)
(box41)--(box52)
(box41)--(box53)
(box41)--(box54)
(box42)--(box54)

(box51)--(box61)
(box52)--(box61)
(box53)--(box61)
(box54)--(box62)

(box61)--(box71)
(box62)--(box71);
\end{tikzpicture}
}
\caption{Construction of $SQ$ when $m=3$. Clearly, $SQ$ is a network of 6 hidden layers each consisting of at most 4 neurons.
For general $m$, one just adds more layers to construct $SQ$ while the number of neurons on each layer is still not exceeding 4.}
\label{figure:construction:SQ:m3}
\end{figure}
\end{proof}

\begin{proposition}\label{proposition:appriximation:product}
For any integer $m\geq 1$, there exists $\xmark_2 \in \mcNN(2m+2, (2,12,\ldots,12,1))$ such that
\begin{eqnarray*} 
0\leq \xmark_2(x, y)\leq 1,\,\,\,\,\bigg|\xmark_2(x, y)-xy\bigg|\leq 4^{-m+1}, \;\;\; \textrm{ for all } x, y\in [0, 1],
\end{eqnarray*}
\end{proposition}
\begin{proof}[{\it Proof of Proposition \ref{proposition:appriximation:product}}]
The proof is a modification of \cite{y17} 
to incorporate normalization.  
Observe that 
$$xy=2\bigg(\frac{x+y}{2}\bigg)^2-\frac{1}{2}x^2-\frac{1}{2}y^2.$$ 
Each of the functions $(x+y)/2, x, y$ can be realized by a network with one hidden layer.
Let $SQ$ denote the network function
in Proposition \ref{proposition:appriximation:square}.
Then we get that for any $0\le x,y\le 1$,
\begin{eqnarray*}
	\bigg|2SQ\bigg(\frac{x+y}{2}\bigg)-\frac{1}{2}SQ(x)-\frac{1}{2}SQ(y)-xy\bigg|\leq 4^{-m},
\end{eqnarray*}
and $$-4^{-m}\leq 2SQ\bigg(\frac{x+y}{2}\bigg)-\frac{1}{2}SQ(x)-\frac{1}{2}SQ(y)\leq 1+ 4^{-m}.$$
Based on the above inequality, we can define
\begin{eqnarray*}
	\xmark_2(x,y)=\frac{2SQ\bigg(\frac{x+y}{2}\bigg)-\frac{1}{2}SQ(x)-\frac{1}{2}SQ(y)+4^{-m}}{1+2\times 4^{-m}},
\end{eqnarray*}
which will be guaranteed to take values in $[0, 1]$. Moreover, for any $0\le x,y\le 1$,
\begin{eqnarray*}
	\bigg|\xmark_2(x,y)-xy\bigg|\leq \frac{4\times4^{-m}}{1+2\times 4^{-m}}\leq 4^{-m+1}.
\end{eqnarray*}
Compared with $SQ$, $\xmark_2$ has two additional hidden layers with two inputs and at most $12$ nodes in each hidden layer;
see Figure \ref{figure:construction:xmark:2}. Proof is complete.
\begin{figure}[ht!]
\centering
 \resizebox {4 in} {2 in} {
\begin{tikzpicture}[>=stealth', shorten >=1pt, auto,
    node distance=2.5cm, scale=1, 
    transform shape, align=center, 
    state/.style={circle, draw, minimum size=0.5cm}, small/.style={ draw}]
\node[small] (x) at (1,3) {$x$};
\node[small] (y) at (1,1) {$y$};

\node[small] (box11) at (3,2) {$x$};
\node[small] (box12) at (3,0) {$y$};
\node[draw, font=\tiny] (box13) at (3,4) {$(x+y)/2$};

\node[state] (box21) at (5,2) {$SQ$};
\node[state] (box22) at (5,0) {$SQ$};
\node[state] (box23) at (5,4) {$SQ$};

\node[small] (box31) at (7,2) {};
\node[small] (box32) at (7,0) {};
\node[small] (box33) at (7,4) {};

\node[small] (box41) at (9,2) {$\xmark_2(x,y)$};
\draw (x)--(box11)
(y)--(box12)
(x)--(box13)
(y)--(box13)

(box11)--(box21)
(box12)--(box22)
(box13)--(box23)

(box21)--(box31)
(box22)--(box32)
(box23)--(box33)

(box31)--(box41)
(box32)--(box41)
(box33)--(box41);
\end{tikzpicture}
}
\caption{Construction of $\xmark_2$. Clearly, $\xmark_2$ has two more hidden layers than $SQ$. On each layer the number of neurons is at most 
three times the number of neurons on each layer of $SQ$, which is 12.}
\label{figure:construction:xmark:2}
\end{figure}
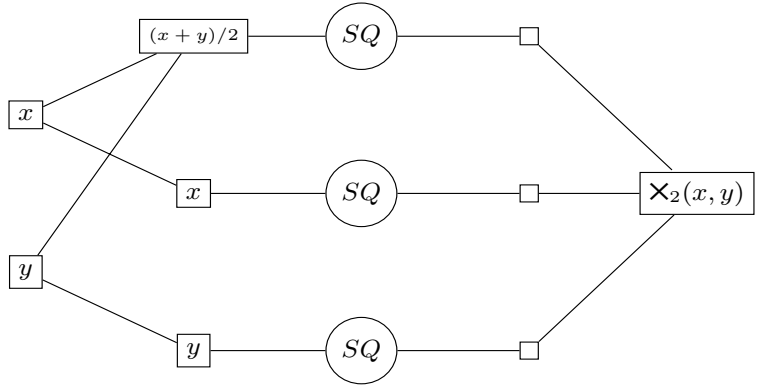
\end{proof}


\begin{proposition}\label{proposition:appriximation:product:k}
For any integers $m\ge1$ and $s\geq 2$, there exists a neural network function $\xmark_s$ with $(s-1)(2m+3)-1$ hidden layers and $10+s$ nodes in each hidden layer such that
for all $x_1, x_2, \ldots, x_s\in [0, 1]$, $0\leq \xmark_s(x_1, x_2, \ldots, x_s)\leq 1$. Moreover, if $|\widetilde{x}_j-x_j|\leq \delta$ with $\widetilde{x}_j \in [0,1]$ for $j=1,2,\ldots,s$, then
\begin{eqnarray*} 
\bigg|\xmark_s(\widetilde{x}_1, \widetilde{x}_2, \ldots, \widetilde{x}_s)-\prod_{j=1}^s x_j\bigg|\leq (s-1)4^{-m+1}+s\delta.
\end{eqnarray*}
\end{proposition}

\begin{proof}[{\it Proof of Proposition \ref{proposition:appriximation:product:k}}]
Let $\delta_m=4^{-m+1}$. Here we only prove the case when $s=3$, and the case for $s>3$ can be proved inductively. First we apply $\xmark_2$ to $x_1, x_2$ and then apply 
$\xmark_2$ to $\xmark_2(x_1, x_2), x_3$. By triangle inequality, we have
\begin{eqnarray*}
	&&\bigg|\xmark_2\bigg(\xmark_2(x_1, x_2), x_3\bigg)-x_1x_2x_3\bigg|\nonumber\\
	&\leq& \bigg|\xmark_2\bigg(\xmark_2(x_1, x_2), x_3\bigg)-\xmark_2(x_1, x_2)x_3\bigg|+\bigg|\xmark_2(x_1, x_2)x_3-x_1x_2x_3\bigg|\nonumber\\
	&\leq& 4^{-m+1}+4^{-m+1}\leq 2\times 4^{-m+1}.
\end{eqnarray*}
In general, let $\xmark_s(x_1, x_2, \ldots, x_s)=\xmark_2\left(\xmark_{s-1}(x_1, x_2,\ldots, x_{s-1}), x_s\right)$ for $s\ge3$.
By induction and triangle inequality, we have
\begin{eqnarray*} 
\bigg|\xmark_s(x_1, x_2, \ldots, x_s)-\prod_{j=1}^s x_j\bigg|\leq (s-1)4^{-m+1}.
\end{eqnarray*}
The desired inequality
follows from the trivial fact that $|\prod_{i=1}^s\widetilde{x}_i-\prod_{i=1}^s{x}_i|\leq s \delta$. Since we apply neural network $\xmark_2$ sequentially $(s-1)$ times and there are $(s-2)$ additional hidden layers to store $\xmark_i(x_1, \ldots, x_i)$ and $x_{i+1}, \ldots, x_s$ for $i=2,\ldots, s-1$ (See Figure \ref{figure:construction:xmark:s}), the total number of hidden layers is $(s-1)(2m+2)+s-2=(s-1)(2m+3)-1$. Moreover, the number of nodes on each hidden layer is at most $12+s-2=10+s$, due to the fact that the first hidden layer has the most number of nodes. Proof is complete.
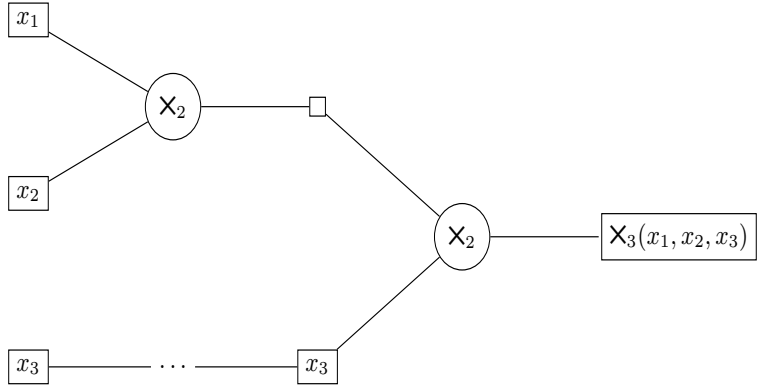
\begin{figure}[ht!]
\centering
 \resizebox {4 in} {2 in} {
\begin{tikzpicture}[>=stealth', shorten >=1pt, auto,
    node distance=2.5cm, scale=1, 
    transform shape, align=center, 
    state/.style={circle, draw, minimum size=0.5cm}, small/.style={ draw}]
\node[small] (x1) at (1,5) {$x_1$};
\node[small] (x2) at (1,3) {$x_2$};
\node[small] (x3) at (1,1) {$x_3$};

\node[state] (box11) at (3,4) {$\xmark_2$};
\node[] (box12) at (3,1) {$\ldots$};

\node[small] (box21) at (5,4) {};
\node[small] (box22) at (5,1) {$x_3$};

\node[state] (box31) at (7,2.5) {$\xmark_2$};

\node[small] (box41) at (10,2.5) {$\xmark_3(x_1, x_2, x_3)$};
\draw (x1)--(box11)
(x2)--(box11)
(x3)--(box12)

(box11)--(box21)
(box12)--(box22)
(box21)--(box31)
(box22)--(box31)

(box31)--(box41);
\end{tikzpicture}
}
\caption{Construction of $\xmark_s$ with $s=3$. $\xmark_3$ links two $\xmark_2$ structures sequentially  and adds one more hidden layer in the mid. The number of neurons on each hidden layer of $\xmark_3$ is at most 1 plus the number of neurons on each hidden layer of $\xmark_2$, which is 13.}
\label{figure:construction:xmark:s}
\end{figure}
\end{proof}

\begin{figure}[ht!]
\centering
\subfigure[]{\includegraphics[width=2 in, height=1.5 in]{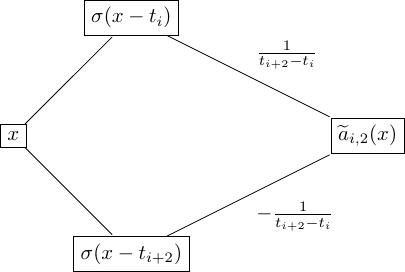}}
\hspace{1cm}
\subfigure[]{\includegraphics[width=2 in, height=1.5 in]{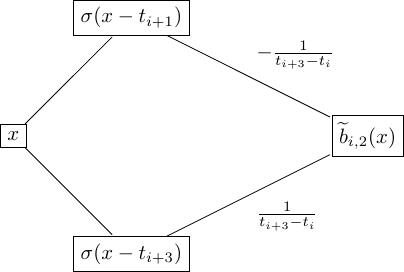}}
\vspace{0.5cm}

\subfigure[]{\includegraphics[width=2 in, height=1.5 in]{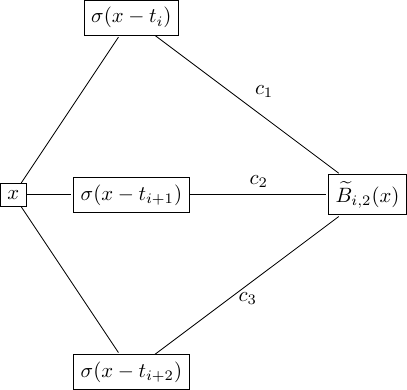}}
\hspace{1cm}
\subfigure[]{\includegraphics[width=2 in, height=1.5 in]{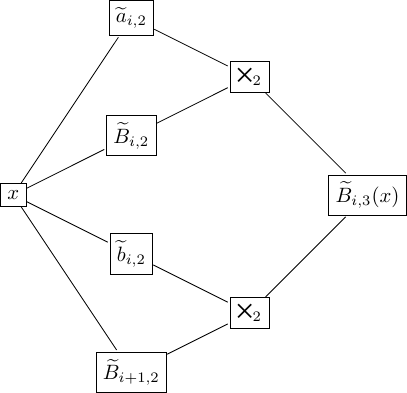}}
\caption{Construction of $\widetilde{B}_{i,3}$ through induction. 
(a) and (b)  demonstrate the architectures of the networks $\widetilde{a}_{i,2}$ and $\widetilde{b}_{i,2}$. 
(c) demonstrates the architecture of the network $\widetilde{B}_{i,2}$ with $c_1, c_2, c_3$ defined in (\ref{eq:definition:c1c2c3}). 
(d) demonstrates the induction relationship between $\widetilde{B}_{i,3}$ and $\widetilde{B}_{i,2}$.
}
\label{figure:tildeBi2:to:tildeBi3}
\end{figure}

Given Proposition \ref{proposition:appriximation:product:k}, we are ready to approximate the $k$th order univariate B-spline basis $B_{i,k}$. Fixing integer $m\geq 1$,
our method is based on the induction formula (\ref{induction:formula}) which allows us to start from approximating $B_{i,2}$.
Specifically, we approximate $B_{i,2}$ by $\widetilde{B}_{i, 2}$ defined as
\begin{eqnarray*}
	\widetilde{B}_{i, 2}(x)=c_1\sigma(x-t_i)+c_2\sigma(x-t_{i+1})+c_3\sigma(x-t_{i+2}),
\end{eqnarray*}
where
\begin{eqnarray}
	c_1=\frac{1}{t_{i+1}-t_i}, \;\;c_2=-\frac{t_{i+2}-t_{i}}{t_{i+2}-t_{i+1}}c_1, \;\;c_3=-(t_{i+2}-t_{i}+1)c_1-(t_{i+2}-t_{i+1}+1)c_2.\label{eq:definition:c1c2c3}
\end{eqnarray}
The piecewise linear function $\widetilde{B}_{i,2}$ is exactly a neural network with one hidden layer consisting of three nodes. 
Suppose that we have
constructed $\widetilde{B}_{i,s}(x)$, a neural network approximation of $B_{i,s}$.
Next we will approximate $B_{i,s+1}$. For $-k+1\leq i\leq M+k-s-1$, we define piecewise linear functions
\begin{eqnarray*}
	\widetilde{a}_{i,s}(x)=\begin{cases}
	0, & \textrm{if } x<t_i\\
	\frac{x-t_i}{t_{i+s}-t_i}, &\textrm{if } t_i\leq x \leq t_{i+s}\\
	1,&\textrm{if } x> t_{i+s}\\
	\end{cases},
	\;\;\;\widetilde{b}_{i,s}(x)=\begin{cases}
	1, & \textrm{if } x<t_{i+1}\\
	\frac{t_{i+s+1}-x}{t_{i+s+1}-t_{i+1}}, &\textrm{if } t_{i+1}\leq x \leq t_{i+s+1}\\
	0, &\textrm{if } x> t_{i+s+1}\\
	\end{cases}.
\end{eqnarray*}
In terms of ReLU activation function, we can rewrite the above as 
$\widetilde{a}_{i,s}(x)=\frac{1}{t_{i+s}-t_i}\sigma(x-t_i)-\frac{1}{t_{i+s}-t_i}\sigma(x-t_{i+s})$ and $\widetilde{b}_{i,s}(x)=-\frac{1}{t_{i+s+1}-t_{i+1}}\sigma(x-t_{i+1})+\frac{1}{t_{i+s+1}-t_i}\sigma(x-t_{i+s+1})+1$, which implies that $\widetilde{a}_{i,s}$ and $\widetilde{b}_{i,s}$
are exactly neural networks with one hidden layer consisting of two nodes (see Figure \ref{figure:tildeBi2:to:tildeBi3}). 
For $i=-k+1,\ldots, M+k-s-2$, we define
\begin{eqnarray*}
	\widetilde{B}_{i, s+1}(x)=\frac{\xmark_2(\widetilde{a}_{i,s}(x), \widetilde{B}_{i,s}(x))+\xmark_2(\widetilde{b}_{i,s}(x), \widetilde{B}_{i+1,s}(x))+2\times 4^{-m+1}+\frac{8^{s}}{7}4^{-m}}{1+4\times 4^{-m+1}+\frac{8^{s}}{14}4^{-m+1}},\; \textrm{ for } x\in[0,1].
\end{eqnarray*}
The `seemingly strange' normalizing constant forces $\widetilde{B}_{i, s+1}(x)$ to take values in $[0,1]$.
We repeat the above steps until we reach the construction of $\widetilde{B}_{i, k}$
(see Figure \ref{figure:tildeBi2:to:tildeBi3} for an illustration of such induction). 
We then approximate  ${B}_{i,k}$ by $\widetilde{B}_{i,k}$. 

Finally, let us count the number of nodes in each hidden layer of $\widetilde{B}_{i,k}$.  Suppose $\widetilde{B}_{i,k}$ has $W_k$ nodes in each hidden layer. Since $\widetilde{B}_{i,2}\in \mcNN(1, (1,3,1))$ for  and $\widetilde{a}_{i,s}, \widetilde{b}_{i,s}\in \mcNN(1, (1,2,1))$ for all $i, s$, we know  $W_2=3$. By Figure \ref{figure:tildeBi2:to:tildeBi3}(d) and Proposition \ref{proposition:appriximation:product}, we show that  $W_3\leq \max\{2\times 12, 2\times(2+W_2)\}\leq 2W_2+28$. By induction, we have that 
\begin{equation}\label{eq:width:btildei}
W_k\leq 2W_{k-1}+28\leq 2^{k-2}(W_2+28)-28\leq  2^{k+3}
\end{equation}
We next approximate the tensor product B-spline basis $D_{\bfi,k}(\bfx)=\prod_{j=1}^dB_{i_j,k}(x_j)$ by
\begin{eqnarray*}
	\widetilde{D}_{\bfi, k}(\bfx)=\xmark_d(\widetilde{B}_{i_1,k}(x_1),\widetilde{B}_{i_2,k}(x_2),\ldots, \widetilde{B}_{i_d,k}(x_d)), \textrm{ for each } \bfi=(i_1,\ldots, i_d) \in \Gamma.
\end{eqnarray*}
Finally, parallelizing $\widetilde{D}_{\bfi, k}(\bfx), \bfi \in \Gamma$ according to (\ref{eq:pilot:estimator}), 
we construct $\widehat{f}_{\textrm{net}}$ as 
\begin{eqnarray}
	\widehat{f}_{\textrm{net}}(\bfx)=\sum_{\bfi\in \Gamma}\widehat{b}_{\bfi}\widetilde{D}_{\bfi, k}(\bfx),\,\,\,\,
	x\in\Omega,\label{eq:optimal:DNN:estimator}
\end{eqnarray}
where the coefficients $\widehat{b}_{\bfi}$'s are obtained in (\ref{LSE:eqn}).

In comparing (\ref{eq:pilot:estimator}) with (\ref{eq:optimal:DNN:estimator}), if 
we can show that ${D}_{\bfi, k}$ and $\widetilde{D}_{\bfi,k}$ are close enough, and $\widehat{b}_{\bfi}$'s are uniformly bounded,
then one can expect that $\widehat{f}_{\text{net}}$ performs similarly to $\widehat{f}_{\textrm{pilot}}$. 
A rich class of statistical results in literature enable us to efficiently analyze $\widehat{f}_{\textrm{pilot}}$.
In the rest of our analysis, we focus on cardinal B-splines for convenience. 

\subsection{Approximation Error to B-spline Basis}
The goal of this subsection is to study the differences between $D_{\bfi,k}$'s and $\widetilde{D}_{\bfi,k}$'s.
Let $\bfi_1, \bfi_2,\ldots, \bfi_q$ be the elements of $\Gamma$, where  $q=(M+k-1)^d$ is the total number of tensor product spline basis functions. For simplicity, we define
\begin{eqnarray*}
 \bfB_k(x)&=&(B_{-k+1,k}(x), B_{-k+2,k}(x),\ldots, B_{M-1,k}(x))^T\in \mathbb{R}^{M-k+1},\\
     \bfD_{k}(\bfx)&=&(D_{\bfi_1, k}(\bfx),D_{\bfi_2, k}(\bfx),\ldots, D_{\bfi_q, k}(\bfx))^T\in \mathbb{R}^{q},\\ 
     \widetilde{\bfB}_k(x)&=&(\widetilde{B}_{-k+1,k}(x), \widetilde{B}_{-k+2,k}(x),\ldots, \widetilde{B}_{M-1,k}(x))^T\in \mathbb{R}^{M-k+1},\\
	\widetilde{\bfD}_k(\bfx)&=&(\widetilde{D}_{\bfi_1,k}(\bfx), \widetilde{D}_{\bfi_2,k}(\bfx),\ldots, \widetilde{D}_{\bfi_q,k}(\bfx))^T\in \mathbb{R}^q.
\end{eqnarray*}
Lemmas \ref{lemma:approximation:b:spline:d:1} and \ref{lemma:approximation:b:spline:d:d} bound the approximation errors of $\widetilde{\bfB}_k(\cdot)$ and $\widetilde{\bfD}_k(\cdot)$. 

\begin{lemma}\label{lemma:approximation:b:spline:d:1}
Given integers $k, M, m\geq 2$ and knots $t_{-k+1}<t_{-k+2}<\ldots< t_0<t_1< \ldots<t_M< t_{M+1} <\ldots< t_{M+k-1}$ such that $t_0=0, t_M=1$,
there exists a $\widetilde{\bfB}_k\in \mcNN(k(2m+3), (1, 2^{k+4}(M+2k), \ldots, 2^{k+4}(M+2k), M+k-1))$
taking values in $[0,1]$, such that
\begin{eqnarray*}
	\sup_{x\in[0,1]}\|\widetilde{\bfB}_k(x)-\bfB_{k}(x)\|_\infty\leq \frac{8^{k}}{14}4^{-m}.
\end{eqnarray*}
\end{lemma}
\begin{proof}[{\it Proof of Lemma \ref{lemma:approximation:b:spline:d:1}}]
First we will approximate $B_{i, 2}$, the linear B-spline, using ReLU neural network. Review that for $i=-k+1,\ldots,M+k-3$,
\begin{eqnarray*}
	B_{i, 2}(x)=\begin{cases}
		\frac{x-t_i}{t_{i+1}-t_i}, & \textrm{if } x\in [t_i, t_{i+1}]\\
		\frac{t_{i+2}-x}{t_{i+2}-t_{i+1}}, & \textrm{if } x\in [t_{i+1}, t_{i+2}]\\
		0, & \textrm{elsewhere}
	\end{cases}.
\end{eqnarray*}
It is easily verified that $B_{i, 2}(x)=c_1\sigma(x-t_i)+c_2\sigma(x-t_{i+1})+c_3\sigma(x-t_{i+2})$, where
\begin{eqnarray*}
	c_1=\frac{1}{t_{i+1}-t_i}, \;\;c_2=-\frac{t_{i+2}-t_{i}}{t_{i+2}-t_{i+1}}c_1, \;\;c_3=-(t_{i+2}-t_{i}+1)c_1-(t_{i+2}-t_{i+1}+1)c_2.
\end{eqnarray*}
This implies that $B_{i, 2}$ is exactly a ReLU neural network (hence, $\widetilde{B}_{i,2}=B_{i,2}$) 
with approximation error $\delta_2=\sup_{x\in [0,1]}|\widetilde{B}_{i,2}(x)-B_{i,2}(x)|=0$ for all $-k+1\leq i \leq M+k-3$. 
Trivially, $B_{i,2}$ takes values in $[0,1]$.

Suppose that we have constructed a neural network approximation $\widetilde{B}_{i,s}$ of $B_{i,s}$ with approximation error 
$\delta_s=\sup_{x\in [0,1]}|\widetilde{B}_{i,s}(x)-B_{i,s}(x)|$. Moreover, $0\leq \widetilde{B}_{i,s}(x) \leq 1$ for all $x\in[0,1]$.

Now we will approximate $B_{i, s+1}$. By definition B-splines, we have
\begin{eqnarray}
	B_{i, s+1}(x)=\frac{x-t_i}{t_{i+s}-t_i}B_{i, s}(x)+\frac{t_{i+s+1}-x}{t_{i+s+1}-t_{i+1}}B_{i+1, s}(x).\label{eq:lemma:approximation:b:spline:d:1:eq1}
\end{eqnarray}
Let us recall the previously defined piecewise linear functions: 
\begin{eqnarray*}
	{a}_{i,s}(x)=\begin{cases}
	0, & \textrm{if } x<t_i\\
	\frac{x-t_i}{t_{i+s}-t_i}, &\textrm{if } t_i\leq x \leq t_{i+s}\\
	0,&\textrm{if } x> t_{i+s}\\
	\end{cases},\;\;\;\widetilde{a}_{i,s}(x)=\begin{cases}
	0, & \textrm{if } x<t_i\\
	\frac{x-t_i}{t_{i+s}-t_i}, &\textrm{if } t_i\leq x \leq t_{i+s}\\
	1,&\textrm{if } x> t_{i+s}\\
	\end{cases}.
\end{eqnarray*}
Notice that
the first term of the right side of (\ref{eq:lemma:approximation:b:spline:d:1:eq1}) is $a_{i,s} B_{i, s}$, 
which can be approximated by $\xmark_2(\widetilde{a}_{i,s}, \widetilde{B}_{i,s})$.
Clearly, $\widetilde{a}_{i,s}(x)=\frac{1}{t_{i+s}-t_i}\sigma(x-t_i)+\sigma(x-t_{i+s})$, which also can be expressed as a ReLU neural network.
Moreover, for any $x\in[0,1]$, it follows by Proposition \ref{proposition:appriximation:product:k} that
\begin{eqnarray}\label{here:1}
	&&\bigg|\xmark_2(\widetilde{a}_{i,s}(x), \widetilde{B}_{i,s}(x))-a_{i,s}(x)B_{i,s}(x)\bigg|\nonumber\\
	&\leq&\bigg|\xmark_2(\widetilde{a}_{i,s}(x), \widetilde{B}_{i,s}(x))-\widetilde{a}_{i,s}(x)\widetilde{B}_{i,s}(x)\bigg|+\bigg|a_{i,s}(x)B_{i,s}(x)-\widetilde{a}_{i,s}(x)\widetilde{B}_{i,s}(x)\bigg|\nonumber\\
	&\leq& 4^{-m+1}+B_{i,s}(x)\bigg|a_{i,s}(x)-\widetilde{a}_{i,s}(x)\bigg|+\widetilde{a}_{i,s}(x)\bigg|B_{i,s}(x)-\widetilde{B}_{i,s}(x)\bigg|\nonumber\\
	&\leq& 4^{-m+1}+0+\delta_s,
\end{eqnarray}
where the last inequality follows by 
the fact that $B_{i, s}$ is supported on $[t_i, t_{i+s}]$. Similarly, let us recall
\begin{eqnarray*}
	{b}_{i,s}(x)=\begin{cases}
	0, & \textrm{if } x<t_{t+1}\\
	\frac{t_{i+s+1}-x}{t_{i+s+1}-t_{i+1}}, &\textrm{if } t_{i+1}\leq x \leq t_{i+s+1}\\
	0,&\textrm{if } x> t_{i+s+1}\\
	\end{cases},\;\;\;\widetilde{b}_{i,s}(x)=\begin{cases}
	1, & \textrm{if } x<t_{i+1}\\
	\frac{t_{i+s+1}-x}{t_{i+s+1}-t_{i+1}}, &\textrm{if } t_{i+1}\leq x \leq t_{i+s+1}\\
	0,&\textrm{if } x> t_{i+s+1}\\
	\end{cases}.
\end{eqnarray*}
Notice that the second term of
the right side of (\ref{eq:lemma:approximation:b:spline:d:1:eq1}) is $b_{i,s} B_{i+1,s}$.
Similar to (\ref{here:1}) we have, for any $x\in[0,1]$,
\begin{eqnarray*}
	&&\bigg|\xmark_2(\widetilde{b}_{i,s}(x), \widetilde{B}_{i+1,s}(x))-b_{i,s}(x)B_{i+1,s}(x)\bigg|\leq 4^{-m+1}+\delta_s.
\end{eqnarray*}
Now let us recursively define
\begin{eqnarray*}
	\widetilde{B}_{i, s+1}(x)=\frac{\xmark_2(\widetilde{a}_{i,s}(x), \widetilde{B}_{i,s}(x))+\xmark_2(\widetilde{b}_{i,s}(x), \widetilde{B}_{i+1,s}(x))+2\times 4^{-m+1}+2\delta_s}{1+4\times 4^{-m+1}+4\delta_s},
\end{eqnarray*}
which is a ReLU neural network taking values in $[0,1]$. It is not difficult to verify that for any $x\in[0,1]$,
\begin{eqnarray*}
\bigg|\widetilde{B}_{i, s+1}(x)-{B}_{i, s+1}(x)\bigg|\leq \frac{8\times 4^{-m+1}+8\delta_s}{1+4\times 4^{-m+1}+4\delta_s}\leq 8\times 4^{-m+1}+8\delta_s.
\end{eqnarray*}
Taking supremum on the left we get $\delta_{s+1}\le 8\times4^{-m+1}+8\delta_s$.
Using $\delta_2=0$, we can conclude $\delta_s \leq \frac{8^{s}}{14}4^{-m}-\frac{32}{7}4^{-m}\leq \frac{8^{s}}{14}4^{-m}$ for $2\leq s \leq k$. 
Deploy $\widetilde{B}_{i, k}$ parallelly to construct the network $\widetilde{\bfB}_k$. 

To count the number hidden layers, we first notice that $\widetilde{B}_{i,2}\in \mcNN(1, (1,3,1))$ for  and $\widetilde{a}_{i,s}, \widetilde{b}_{i,s}\in \mcNN(1, (1,2,1))$ for all $i, s$ by its construction right below Proposition \ref{proposition:appriximation:product:k}. Moreover, from $\widetilde{B}_{i,2}$ to $\widetilde{B}_{i, k}$, we used the network $\xmark_2$ $k-2$ times. Therefore, by Proposition \ref{proposition:appriximation:product},  the number of hidden layers is at most $(2m+2)(k-2)+k-2+1$, which is bounded by $(2m+3)k$. Since in each hidden layer, at most we have $M+2k-3$ different $\widetilde{B}_{i,s}$'s, $\widetilde{a}_{i,s}$'s  and $\widetilde{b}_{i,s}$'s for $s=2,\ldots, k$. So by (\ref{eq:width:btildei}), at most, we have $(2^{k+3}+4)(M+2k)\leq 2^{k+4}(M+2k)$ nodes in each hidden layer. The proof is complete.
\end{proof}


\begin{lemma}\label{lemma:approximation:b:spline:d:d}
Given integers $k, M, m\geq 2$ and knots $t_{-k+1}<t_{-k+2}<\ldots< t_0<t_1< \ldots<t_M< t_{M+1} <\ldots< t_{M+k-1}$ with 
$t_0=0, t_M=1$, there exists a
$\widetilde{\bfD}_k\in \mcNN((2m+3)(k+d-1), (d, 2^{k+4}d(M+2k)^d, \ldots,  2^{k+4}d(M+2k)^d, (M+k-1)^d))$ such that
\begin{eqnarray*}
	\sup_{\bfx \in \Omega}\bigg\|\widetilde{\bfD}_k(\bfx)-\bfD_{k}(\bfx)\bigg\|_{\infty}\leq [4(d-1)+8^k]4^{-m}.
\end{eqnarray*}
Furthermore, each element of $\widetilde{\bfD}_k$ is in $[0,1]$.
\end{lemma}
\begin{proof}[{\it Proof of Lemma \ref{lemma:approximation:b:spline:d:d}}]
Let $\widetilde{\bfB}_k(x_1), \widetilde{\bfB}_k(x_1), \ldots, \widetilde{\bfB}_k(x_d)$ be the neural networks provided 
in Lemma \ref{lemma:approximation:b:spline:d:1}, which satisfy $|\widetilde{B}_{i, k}(x)-{B}_{i, k}(x)|\leq \delta_m$, where
$\delta_m=8^k4^{-m}/14$. For each $(i_1, i_2, \ldots, i_d)\in \{-k+1, -k+2, \ldots, 1,2, \ldots, M-1\}^d$, we apply the
product network $\xmark_d$ given in Proposition \ref{proposition:appriximation:product:k} to 
$(\widetilde{B}_{i_1,k}(x_1), \widetilde{B}_{i_2,k}(x_2),\ldots, \widetilde{B}_{i_d,k}(x_d))$. According to Proposition \ref{proposition:appriximation:product:k}, we have
\begin{eqnarray*}
	\bigg|\xmark_d(\widetilde{B}_{i_1,k}(x_1), \widetilde{B}_{i_2,k}(x_2),\ldots, \widetilde{B}_{i_d,k}(x_d))-\prod_{j=1}^d B_{i_j,k}(x_j)\bigg|&\leq& (d-1)4^{-m+1}+d\delta_m\\
	&\leq& [4(d-1)+8^k]4^{-m}.
\end{eqnarray*}
Now we deploy $\xmark_d(\widetilde{B}_{i_1,k}(x_1), \widetilde{B}_{i_2,k}(x_2),\ldots, \widetilde{B}_{i_d,k}(x_d))$ parallelly to construct the network $\widetilde{\bfD}_k$. Since we apply neural network $X_d$ to output of $\widetilde{\bfB}_k$, so the total number of hidden layers is at most $k(2m+3)+1+(d-1)(2m+3)-1\leq (2m+3)(d+k)$. Since we parallelly apply $q=(M+k-1)^d$ product networks $\xmark_d$,  the number nodes in each hidden layer is bounded $(10+d)q$, which is further bounded by $d2^{k+4}(M+2k)^d$. This completes the proof.
\end{proof}
In \cite{ES19}, the authors compare neural network methods with multivariate adaptive regression splines (MARS)
by showing that any function expressed by MARS can be approximated by a sparse ReLU neural network with an arbitrarily small error.
In contrast, Lemma \ref{lemma:approximation:b:spline:d:1} provides a quantitative error bound (in terms of network architecture) 
for fully connected ReLU neural network approximation of the spline basis.
Soon after our work, \cite{KKL19} independently obtain a relevant result
about a quantitative connection between MARS and sparse neural network under smooth activation function.

To end this subsection, let us calculate the number of hidden layers and number of nodes in each hidden layer for $\widehat{f}_{\textrm{net}}$ defined in (\ref{eq:optimal:DNN:estimator}).
Notice that to construct  $\widehat{f}_{\textrm{net}}$, we only need to add one more hidden layer to aggregate  $\widetilde{D}_{\bfi, k}(\bfx)$ and the coefficients $\widehat{b}_{\bfi}$. As a consequence, for any integers $k, M, m\geq 2$, we can construct a network $\widehat{f}_{\textrm{net}}$ such that
\begin{equation}\label{eq:counting:fnet}
\widehat{f}_{\textrm{net}}\in \mcF(L, \bfp(T)),\quad \textrm{with $L=(2m+3)(k+d)+1$ and $T=2^{k+4}d(M+2k)^d$}.
\end{equation}
By Proposition \ref{proposition:appriximation:product:k}, we expect $\widehat{f}_{\textrm{net}}\approx \widehat{f}_{\textrm{pilot}}$ when $m \to \infty$ (or equivalently $L\to \infty$).

\subsection{Asymptotic Properties of the Pilot Estimator}\label{sec:aymptpt:pilot}
In this subsection, we study the convergence rate of the pilot estimator in (\ref{eq:pilot:estimator}) and the bound of coefficients in (\ref{LSE:eqn}). Let us define $\Phi=(\bfD_k(\bfX_1),\ldots, \bfD_k(\bfX_n))^T\in \mathbb{R}^{n\times q}$ and $\bfY=(Y_1,\ldots Y_n)^T$. Therefore, the coefficients in (\ref{LSE:eqn}) can be expressed as  $\widehat{C}=(\Phi^T\Phi)^{-1}\Phi^T\bfY$, where the invertibility of the matrix $\Phi^T\Phi$ is guaranteed by Lemma \ref{lemma:empirical:eigen:value} below. Moreover,  we denote $\Theta_n=\{g(\bfx)| g(\bfx)=V^T\bfD_k(\bfx) \textrm{ for } V\in \mathbb{R}^q\}$ as the linear space spanned by the tenor product B-spline basis $\bfD_k$'s. An additional assumption is to obtain the desired results, which is stated as follows.
\begin{Assumption}\label{Assumption:A1}
\label{A1:c} 
The knots $\{t_i, i=-k+1,\ldots, M+k-1\}$ have constant separation $h=M^{-1}$. In the theoretical analysis, we require $M\to \infty$ and $h\to 0$.
\end{Assumption}

\begin{remark}
Assumption \ref{Assumption:A1} can be relaxed to $\max_{i}(t_{i+1}-t_{i})/\min_{i}(t_{i+1}-t_{i})\leq c$ for some constant $c>0$,
under which one needs to redefine the separation
$h=\max_{i}(t_{i+1}-t_{i})$. Results in this section continue to hold.
This is a standard assumption for $B$-spline literature; see \cite{h98}.
\end{remark}

Based on Assumption \ref{Assumption:A1}, we can delivery some preliminary lemmas about the B-spline basis. In particular, Lemma \ref{lemma:spline:approximation} quantifies the approximation error of splines; Lemma \ref{lemma:equivalence:empirical:population:norm} indicates the equivalence of the norms $\|\cdot\|_n$ and $\|\cdot\|_{L^2}$; Lemma \ref{lemma:eigen:value:design:matrix} studies the upper and lower bounds of the eigenvalues for the tensor product B-spline basis matrix.

\begin{lemma}\label{lemma:spline:approximation}
For any $f\in \Lambda^\beta(F, \Omega)$,  suppose that Assumption \ref{Assumption:A1} is satisfied with some integer $k\geq \beta$.  There exists a real sequence $c_\bfi$ such that
	$\sup_{\bfx \in \Omega}\bigg|\sum_{\bfi\in \Gamma}c_\bfi D_{\bfi,k}(\bfx)-f(\bfx)\bigg|\leq A_f h^\beta$ and $|c_\bfi|\leq A_f$ for all $\bfi \in \Gamma$. Here
 $A_f>0$ is a constant only relying on $F, \beta, k$ and $\|f\|_{\sup}$. Moreover, it holds that $sup_{f\in \Lambda^\beta(F, \Omega)}A_f<\infty$, where the upper bound only depends on $F$, $\beta$ and $k$.
\end{lemma}
The proof of Lemma \ref{lemma:spline:approximation} requires borrowing some definition from \cite{gkkw06}. Hence, we defer its proof to the Appendix.

\begin{lemma}\label{lemma:equivalence:empirical:population:norm}
Suppose Assumptions \ref{A0} and \ref{Assumption:A1} hold with some integer $k\geq \max(\beta, 2)$. Moreover, if the sequence $h$ in Assumption \ref{Assumption:A1} satisfies $h=o(1)$ and $\log(h^{-1})=o(nh^d)$, then 
\begin{eqnarray*}
	\sup_{g\in \Theta_n}\bigg|\frac{\|g\|_n^2}{\|g\|_{L^2}^2}-1\bigg|=o_P(1).
\end{eqnarray*}
\end{lemma}
\begin{proof}[Proof of Lemma  \ref{lemma:equivalence:empirical:population:norm}]
This is Lemma 2.3 in \cite{h03}.
\end{proof}



\begin{lemma}\label{lemma:eigen:value:design:matrix}
Suppose Assumptions \ref{A0} and \ref{Assumption:A1} hold with some integer $k\geq \max(\beta, 2)$. Let us define matrix $\bfB=\int_\Omega \bfD_k(\bfx)\bfD_k^T(\bfx)Q(\bfx)d\bfx$. Then the eigenvalues of $\bfB$ satisfy that
\begin{eqnarray*}
	a_1h^d\leq \lambda_{\min}(\bfB)\leq  \lambda_{\max}(\bfB)\leq a_2h^d,
\end{eqnarray*}
where $0<a_1\leq a_2<\infty$ are constants relying on $k$ and density function $Q$. 
\end{lemma}
\begin{proof}[{\it Proof of Lemma \ref{lemma:eigen:value:design:matrix}}]
It follows from \cite[page 155]{d78} that for some constant $\lambda>1$ depending on $k$, we have
\begin{eqnarray*}
	\lambda^{-1}h\leq  \lambda_{\min}\left(\int_0^1\bfB_k(x)\bfB_k^T(x)dx\right)\leq 
	\lambda_{\max}\left(\int_0^1\bfB_k(x)\bfB_k^T(x)dx\right)\leq\lambda h.
\end{eqnarray*}
Notice that $\bfD_k(\bfx)\bfD_k^T(\bfx)=\otimes_{j=1}^d\bfB_k(x_j)\bfB_k^T(x_j)$ for any $\bfx=(x_1, x_2, \ldots, x_d)^T \in [0,1]^d$. Here $\otimes$  is the outer product operator.
It follows that
\begin{eqnarray*}
	\int_{[0,1]^d}\bfD_k(\bfx)\bfD_k^T(\bfx)d\bfx=\otimes_{j=1}^d\int_0^1\bfB_k(x_j)\bfB_k^T(x_j)dx_j.
\end{eqnarray*}
By the property of tensor product of matrix, we have
\begin{eqnarray*}
	\lambda_{\max}\bigg(\int_{[0,1]^d}\bfD_k(\bfx)\bfD_k^T(\bfx)d\bfx\bigg)&=&\lambda^d_{\max}\bigg(\int_0^1\bfB_k(x)\bfB_k^T(x)dx\bigg)\leq \lambda^dh^d,\\
\lambda_{\min}\bigg(\int_{[0,1]^d}\bfD_k(\bfx)\bfD_k^T(\bfx)d\bfx\bigg)&=&\lambda^d_{\min}\bigg(\int_0^1\bfB_k(x)\bfB_k^T(x)dx\bigg)\geq \lambda^{-d}h^d.\\	
\end{eqnarray*}
By Assumption \ref{A0}, there exists a constant $c>1$ such that $c^{-1}\int g(\bfx)d\bfx\leq \int g(\bfx)Q(\bfx)d\bfx\leq c
\int g(\bfx)d\bfx$ for any integrable $g$, which leads to 
\begin{eqnarray*}
	C^T\bigg(\int_{[0,1]^d}\bfD_k(\bfx)\bfD_k^T(\bfx)Q(\bfx)d\bfx\bigg)C^T&=&\int_{[0,1]^d}|C^T\bfD_k(\bfx)|^2Q(\bfx)d\bfx\\
	&\leq& c\int_{[0,1]^d}|C^T\bfD_k(\bfx)|^2d\bfx\\
	&\leq& c\lambda^dh^d, \textrm{ for all } C \in \mathbb{R}^q.
\end{eqnarray*}
Therefore, we have $\lambda_{\max}(\bfB)\leq a_2h^d$ with $a_2=c\lambda^d$. Similarly, we can show that the lower bound is valid with $a_1=a_2^{-1}$. 
Proof is complete.
\end{proof}

To proceed, we need to define the following event
\begin{eqnarray}
\Omega_n&=&\left\{a_1h^d/2\leq \lambda_{\min}(n^{-1}\Phi^T \Phi)\leq  \lambda_{\max}(n^{-1}\Phi^T \Phi)\leq 2a_2h^d\right\}\nonumber\\
&&\cap \left\{\|g\|_{L^2}^2/2\leq \|g\|_n^2\leq 2\|g\|_{L^2}^2, \textrm{ for all } g\in \Theta_n\right\},\label{eq:definition:Omegan}
\end{eqnarray}
where $a_1, a_2$ are the constants introduced in Lemma \ref{lemma:eigen:value:design:matrix}. The following lemma reveals the probability of $\Omega_n$ approaches one as $n$ diverges, which suggests we can focus our analysis on the event $\Omega_n$.
\begin{lemma}\label{lemma:empirical:eigen:value}
Suppose Assumptions \ref{A0} and \ref{Assumption:A1} hold with some integer $k\geq \max(\beta, 2)$. Moreover, if the sequence $h$ in Assumption \ref{Assumption:A1} satisfies $h=o(1)$ and $\log(h^{-1})=o(nh^d)$, then it follows that $\lim_{n\to \infty}P(\Omega_n)=1.$
\end{lemma}
\begin{proof}[{\it Proof of Lemma \ref{lemma:empirical:eigen:value}}]
Notice that $n^{-1}\Phi^T\Phi=\sum_{i=1}^n \bfD_k(\bfX_i)\bfD_k^T(\bfX_i)/n$. Let $\widehat{\bfB}=n^{-1}\Phi^T\Phi$ and $\bfB=\int_\Omega \bfD_k(\bfx)\bfD_k^T(\bfx)Q(\bfx)d\bfx$.
It follows from Lemma \ref{lemma:equivalence:empirical:population:norm} that
\begin{eqnarray*}
\sup_{u \in \mathbb{R}^q}\bigg|\frac{u^T\widehat{\bfB}u}{u^T\bfB u}-1\bigg|
&=&\sup_{u \in \mathbb{R}^q}\bigg|\frac{\sum_{i=1}^n |u^T\bfD_k(\bfX_i)|^2/n}{\int_\Omega |u^T\bfD_k(\bfx)|^2Q(\bfx)d\bfx}-1\bigg|\\
&=&\sup_{g\in \Theta_n}\bigg|\frac{\|g\|_n^2}{\|g\|_{L^2}^2}-1\bigg|=o_P(1),
\end{eqnarray*}
So the event
\begin{eqnarray*}
	K_n=\bigg\{\sup_{u \in \mathbb{R}^q}\bigg|\frac{u^T\widehat{\bfB}u}{u^T\bfB u}-1\bigg|\leq \min(a_2, a_1/2)\bigg\}
\end{eqnarray*}
has probability approaching one. By Lemma \ref{lemma:eigen:value:design:matrix}, on the event $K_n$, it follows that
\begin{eqnarray*}
	\sup_{\|u\|_2=1}|{u^T\widehat{\bfB}u}|&\leq& \sup_{\|u\|_2=1}|{u^T\bfB u}|+ \sup_{\|u\|_2=1}|{u^T\widehat{\bfB}u}-{u^T\bfB u}|\\
	&\leq& a_2h^d+ \sup_{\|u\|_2=1}\bigg|\frac{u^T\widehat{\bfB}u}{u^T\bfB u}-1\bigg| \sup_{\|u\|_2=1}|{u^T\bfB u}|\\
	&\leq& 2a_2h^d.
\end{eqnarray*}
Similarly, we can show $\inf_{\|u\|_2=1}|{u^TAu}|\geq a_1h^d/2$, on the event $K_n$. Above argument and Lemma \ref{lemma:equivalence:empirical:population:norm}  together complete the proof.
\end{proof}

Based on the above lemmas, we are ready to provide the main result in this subsection, which provides the convergence rate of the pilot estimator and the bound of $\widehat{C}$.
\begin{lemma}\label{lemma:rate:of:convergence:sieve}
Suppose Assumptions \ref{A0} and \ref{Assumption:A1} hold with some integer $k\geq \max(\beta, 2)$. Moreover, if the sequence $h$ in Assumption \ref{Assumption:A1} satisfies $h=o(1)$ and $\log(h^{-1})=o(nh^d)$, then it follows that
\begin{eqnarray*}
	\sup_{f_0\in\Lambda^\beta(F,\Omega)}
	\ev_{f_0}\bigg\{\|\widehat{f}_{\textrm{pilot}}-f_0\|_{L^2}^2\bigg|\mathbb{X}\bigg\}=O_P\bigg(h^{2\beta}+\frac{1}{nh^d}\bigg)
\end{eqnarray*}
and
\[
\sup_{f_0\in\Lambda^\beta(F,\Omega)}
\ev_{f_0}\left(\widehat{C}^T\widehat{C}\big|\mathbb{X}\right)=O_P(h^{-d}).
\]
\end{lemma}
\begin{proof}[{\it Proof of Lemma \ref{lemma:rate:of:convergence:sieve}}]
For any $f_0\in\Lambda^\beta(F, \Omega)$, let $\mathbf{f}_0=(f_0(\bfX_1),\ldots,f_0(\bfX_n))^T$.
Also let 
$\bm{\epsilon}=(\epsilon_1,\ldots,\epsilon_n)^T$,
$\widehat{\bff}_{\textrm{pilot}}=(\widehat{f}_{\textrm{pilot}}(\bfX_1),
\ldots,\widehat{f}_{\textrm{pilot}}(\bfX_n))^T$. 
According to Lemma \ref{lemma:spline:approximation} and by $k\geq \beta$, 
there exists a $C=(c_1, c_2,\ldots, c_q)^T\in \mathbb{R}^q$ 
such that for any $\bfx\in\Omega$,
$|C^T\bfD_k(\bfx)-f_0(\bfx)|\leq A_{f_0}h^\beta$.  For simplicity, we further define $f^*(\bfx)=C^T\bfD_k(\bfx)$ and $\bff^*=(f^*(\bfX_1),\ldots, f^*(\bfX_n))^\top$.

Notice that on the event $\Omega_n$, $\Phi^T\Phi$ is invertible. The  least square algorithm (\ref{LSE:eqn}) implies the following holds on event $\Omega_n$:
\begin{eqnarray}
	\widehat{\bff}_{\textrm{pilot}}=\Phi(\Phi^T \Phi)^{-1}\Phi^T \bfY&=&\Phi(\Phi^T \Phi)^{-1}\Phi^T (\Phi C+\bfE+\bm{\epsilon})\nonumber\\
	&=&\Phi C+\Phi(\Phi^T \Phi)^{-1}\Phi^T \bfE+\Phi(\Phi^T \Phi)^{-1}\Phi^T \bfepsilon\nonumber\\
	&=&\bff^*+\Phi(\Phi^T \Phi)^{-1}\Phi^T \bfE+\Phi(\Phi^T \Phi)^{-1}\Phi^T\bfepsilon, \label{eq:lemma:rate:of:convergence:sieve:eq0}
\end{eqnarray}
where $\bfE=(E_1, E_2, \ldots, E_n)^T\in \mathbb{R}^n$ with $E_i=f_0(\bfX_i)-C^T\bfD_k(\bfX_i)=f_0(\bfX_i)-f^*(\bfX_i)$. Furthermore, the above equation and Lemma \ref{lemma:spline:approximation} together imply that
 \begin{eqnarray*}
	\|\widehat{f}_{\textrm{pilot}}-f^*\|_n^2&=&\frac{1}{n}(\widehat{\bff}_{\textrm{pilot}}-\bff^*)^T(\widehat{\bff}_{\textrm{pilot}}-\bff^*)\nonumber\\
	&\leq& \frac{2}{n}\bfE^T\Phi(\Phi^T \Phi)^{-1}\Phi^T\bfE+\frac{2}{n}\bfepsilon\Phi(\Phi^T \Phi)^{-1}\Phi^T \bfepsilon\\
	&\leq& 2A_{f_0}^2h^{2\beta}+\frac{2}{n}\bfepsilon^T\Phi(\Phi^T \Phi)^{-1}\Phi^T \bfepsilon.
\end{eqnarray*}
By the fact that $\widehat{f}_{\textrm{pilot}}-f^*\in \Theta_n$, it holds on event $\Omega_n$ that 
\begin{eqnarray*}
\|\widehat{f}_{\textrm{pilot}}-f^*\|_{L^2}^2\leq 2\|\widehat{f}_{\textrm{pilot}}-f^*\|_n^2.
\end{eqnarray*}
and
\begin{eqnarray*}
	\ev_{f_0}\bigg(\bfepsilon^T\Phi(\Phi^T \Phi)^{-1}\Phi^T \bfepsilon\bigg|\mathbb{X}\bigg)=\textrm{Tr}\bigg(\Phi(\Phi^T \Phi)^{-1}\Phi^T \bigg)=q=(M+k-1)^d\leq 2^dh^{-d},
\end{eqnarray*}
which further implies that
\begin{eqnarray}
\ev_{f_0}\bigg(\|\widehat{f}_{\textrm{pilot}}-f^*\|_{L^2}^2\bigg|\mathbb{X}\bigg)\leq 2\ev_{f_0}\bigg(\|\widehat{f}_{\textrm{pilot}}-f^*\|_n^2\bigg|\mathbb{X}\bigg)\leq 4A_{f_0}^2h^{2\beta}+\frac{2^{d+2}}{nh^d}.\label{eq:lemma:rate:of:convergence:sieve:eq1}
\end{eqnarray}
By simple algebra, the above inequality implies that
\begin{eqnarray*}
	\ev_{f_0}\bigg(\|\widehat{f}_{\textrm{pilot}}-f_0\|_{L^2}^2\bigg|\mathbb{X}\bigg)&\leq&2\ev_{f_0}\bigg(\|\widehat{f}_{\textrm{pilot}}-f^*\|_{L^2}^2\bigg|\mathbb{X}\bigg)+2\ev_{f_0}\bigg(\|f^*-f_0\|_{L^2}^2\bigg|\mathbb{X}\bigg)\\
	&\leq&8A_{f_0}^2h^{2\beta}+\frac{2^{d+3}}{nh^d}+2\|f^*-f_0\|_{\sup}^2\\
	&=&\frac{2^{d+3}}{nh^d}+10A_{f_0}^2h^{2\beta}, \textrm{ uniformly for all } f_0\in\Lambda^\beta(F,\Omega).
\end{eqnarray*}
Finally, the first statement follows by the uniform boundedness of $A_{f_0}$ over $f_0\in\Lambda^\beta(F,\Omega)$ in Lemma \ref{lemma:spline:approximation}
and $\pr(\Omega_n)\to 1$ in Lemma \ref{lemma:empirical:eigen:value}. 

Let us prove the second statement. According to Lemma \ref{lemma:eigen:value:design:matrix}, it follows that
\begin{eqnarray*}
	\|\widehat{f}-g^*\|^2_{L^2}&=&(\widehat{C}-C)^T\int\bfD_k(\bfx)\bfD_k^T(\bfx)Q(\bfx)d\bfx(\widehat{C}-C)\\
	&\geq& a_1h^d(\widehat{C}-C)^T(\widehat{C}-C),
\end{eqnarray*}
where $a_1>0$ is the constant in Lemma \ref{lemma:eigen:value:design:matrix}. Taking conditional expectation and by (\ref{eq:lemma:rate:of:convergence:sieve:eq1}), on event $\Omega_n$,  we have
\begin{eqnarray*}
	\ev_{f_0}\bigg((\widehat{C}-C)^T(\widehat{C}-C)\bigg|\mathbb{X}\bigg)&\leq& \ev_{f_0}\bigg(\|\widehat{f}_{\textrm{pilot}}-f^*\|_{L^2}^2\bigg|\mathbb{X}\bigg)\\
	&\leq&a_1^{-1}2^{d+2}A_{f_0}^2\bigg(h^{2\beta-d}+\frac{1}{nh^{2d}}\bigg), \textrm{ uniformly for all } f_0\in\Lambda^\beta(F,\Omega),
\end{eqnarray*}
which further leads to 
\begin{eqnarray*}
	\ev_{f_0}\bigg(\widehat{C}^T\widehat{C}\bigg|\mathbb{X}\bigg)&\leq&2\ev_{f_0}\bigg((\widehat{C}-C)^T(\widehat{C}-C)\bigg|\mathbb{X}\bigg)+2C^TC\\
	&\leq&a_1^{-1}2^{d+3}A_{f_0}^2\bigg(h^{2\beta-d}+\frac{1}{nh^{2d}}\bigg)+2qA_{f_0}^2\\
	&\leq&a_1^{-1}2^{d+3}A_{f_0}^2\bigg(h^{2\beta-d}+\frac{1}{nh^{2d}}\bigg)+2^{d+1}h^{-d}A_{f_0}^2\\
	&\leq&a_1^{-1}2^{d+3}A_{f_0}^2\bigg(h^{2\beta-d}+\frac{1}{nh^{2d}}+h^{-d}\bigg)\\
	&\leq&a_1^{-1}2^{d+4}A_{f_0}^2h^{-d}, \textrm{ uniformly for all } f_0\in\Lambda^\beta(F,\Omega),
\end{eqnarray*}
where the last inequality holds by the fact $h^{2\beta}+n^{-1}h^{-d}=o(1)$. 
Finally, the second statement follows by the uniform boundedness of $A_{f_0}$ over $f_0\in\Lambda^\beta(F,\Omega)$ in Lemma \ref{lemma:spline:approximation}
and $\pr(\Omega_n)\to 1$ in Lemma \ref{lemma:empirical:eigen:value}. 
Proof is complete.
\end{proof}

\subsection{Approximation Error to the Pilot Estimator}
The following Lemma \ref{thm:approximation:sieve:DNN}
is the main technical result of this paper,
based on which Theorem \ref{thm:neural:estimator:rate:of:convergence:main:text} will be proved.
\begin{lemma}\label{thm:approximation:sieve:DNN}
Suppose Assumptions \ref{A0} and \ref{Assumption:A1} hold with some integer $k\geq \max(\beta, 2)$ and diverging sequence $M$.  Let $m$ be diverging with respect to sample size $n$ and  $F>0$ be a fixed constant. If $M^d\log(M)=o(n)$, then the network function  $\widehat{f}_{\textrm{net}}\in \mathcal{F}(L,\bfp(T))$, with $L=(2m+3)(k+d)+1$ and $T=2^{k+4}d(M+2k)^d$ satisfies
\begin{eqnarray*}
	\sup_{f_0\in\Lambda^\beta(F, \Omega)}\ev_{f_0}\left\{\sup_{\bfx \in \Omega}|\widehat{f}_{\text{net}}(\bfx)-\widehat{f}_{\textrm{pilot}}(\bfx)|^2
	\bigg|\mathbb{X}\right\}=O_P(M^{2d}4^{-2m}).
\end{eqnarray*}
Here the $O_P$ is in the sense of diverging $m, M, n$. 
\end{lemma}
\begin{proof}[{\it Proof of Lemma \ref{thm:approximation:sieve:DNN}}]
By the notation in the proof of Lemma \ref{lemma:rate:of:convergence:sieve} and (\ref{eq:lemma:rate:of:convergence:sieve:eq0}), we have
\begin{eqnarray*}
	\widehat{\bff}_{\textrm{pilot}}=\Phi(\Phi^T \Phi)^{-1}\Phi^T \bfY&=&\Phi(\Phi^T \Phi)^{-1}\Phi^T (\Phi C+\bfE+\bm{\epsilon})\\
	&=&\Phi C+\Phi(\Phi^T \Phi)^{-1}\Phi^T \bfE+\Phi(\Phi^T \Phi)^{-1}\Phi^T \bfepsilon\\
	&=&\bff_0-(I-\Phi(\Phi^T \Phi)^{-1}\Phi^T)\bfE+\Phi(\Phi^T \Phi)^{-1}\Phi^T \bfepsilon.
\end{eqnarray*}
It follows from (\ref{eq:pilot:estimator}),  (\ref{LSE:eqn}) and (\ref{eq:optimal:DNN:estimator}) that
$\widehat{f}_{\textrm{pilot}}(\bfx)=\widehat{C}^T\bfD_k(\bfx)$ and $\widehat{f}_{\textrm{net}}(\bfx)=\widehat{C}^T\widetilde{\bfD}_k(\bfx)$.
Therefore, for any $\bfx\in\Omega$, we have
\begin{eqnarray*}
	|\widehat{f}_{\textrm{pilot}}(\bfx)-\widehat{f}_{\textrm{net}}(\bfx)|^2&=&\big\|\widehat{C}^T\left(\bfD_k(\bfx)-\widetilde{\bfD}_k(\bfx)\right)\big\|_2^2\\
	&=& \widehat{C}^T\widehat{C}\left(\bfD_k(\bfx)-\widetilde{\bfD}_k(\bfx)\right)^T\left(\bfD_k(\bfx)-\widetilde{\bfD}_k(\bfx)\right)\\
	&\leq&q\widehat{C}^T\widehat{C} \sup_{\bfx \in [0,1]^d}\big\|\bfD_k(\bfx)-\widetilde{\bfD}_k(\bfx)\big\|_{\infty}^2
	\le q\widehat{C}^T\widehat{C}[4(d-1)+8^k]^24^{-2m},
\end{eqnarray*}
where the last inequality follows from Lemma \ref{lemma:approximation:b:spline:d:d}.
Following Lemma \ref{lemma:rate:of:convergence:sieve} and the fact $q=|\Gamma|=(M+k-1)^d\asymp h^{-d}$, we have
\begin{eqnarray}
	\sup_{f_0\in\Lambda^\beta(F,\Omega)}
	\ev_{f_0}\bigg(\sup_{\bfx \in \Omega}|\widehat{f}_{\textrm{pilot}}(\bfx)-\widehat{f}_{\textrm{net}}(\bfx)|^2 \bigg|\mathbb{X}\bigg)&\leq&
	q[4(k-1)+8k]^2 4^{-2m}\sup_{f_0\in\Lambda(F,\Omega)}\ev\left(\widehat{C}^T\widehat{C}\big|\mathbb{X}\right)\nonumber\\
	&=&O_P(h^{-2d}4^{-2m}),\label{eq:thm:approximation:sieve:DNN:eq1}
\end{eqnarray}
which completes the proof by noticing that $M\asymp h^{-1}$.
\end{proof}
To the end of this section, let us complete the proof of Theorem \ref{thm:neural:estimator:rate:of:convergence:main:text}.  Combining Lemmas \ref{lemma:rate:of:convergence:sieve} and \ref{thm:approximation:sieve:DNN}, we have
\begin{eqnarray*}
	&&\inf_{\widehat{f}\in \mcF(L, \bfp(T))}\sup_{f_0\in\Lambda^\beta(F,\Omega)}
	\ev_{f_0}\bigg(\|\widehat{f}-f_0\|_{L^2}^2|\mathbb{X}\bigg)\\
	&\leq&\sup_{f_0\in\Lambda^\beta(F,\Omega)}
	\ev_{f_0}\bigg(\|\widehat{f}_{\textrm{net}}-f_0\|_{L^2}^2|\mathbb{X}\bigg)\\
	&\leq&2\sup_{f_0\in\Lambda^\beta(F,\Omega)}
	\ev_{f_0}\bigg(\|\widehat{f}_{\textrm{net}}-\widehat{f}_{\textrm{pilot}}\|_{L^2}^2|\mathbb{X}\bigg)+2\sup_{f_0\in\Lambda^\beta(F,\Omega)}
	\ev_{f_0}\bigg(\|\widehat{f}_{\textrm{pilot}}-f_0\|_{L^2}^2|\mathbb{X}\bigg)\\
	&=&O_P\bigg(M^{-2\beta}+\frac{M^d}{n}\bigg)+O_P(M^{2d}4^{-2m})\\
	&=&O_P\bigg(T^{-\frac{2\beta}{d}}+\frac{T}{n}+T^24^{-\frac{L}{k+d}}\bigg)
\end{eqnarray*}
where the fact that $M\asymp h^{-1}$, $L=(2m+3)(k+d)+1$ and $T=2^{k+4}d(M+2k)^d$ is used. We would like to comment that Theorem \ref{thm:neural:estimator:rate:of:convergence:main:text} does not rely on Assumption \ref{Assumption:A1}, as we only need such $\widehat{f}_{\textrm{net}}$ exists. 


\section{Asymptotic Distribution and Optimal Testing}\label{sec:byproducts}
In this subsection,  
we derive the asymptotic distribution for $\widehat{f}_{\textrm{net}}$ and a 
corresponding hypothesis testing procedure.  Let us recall that the network function constructed in (\ref{eq:counting:fnet}) satisfies 
\begin{eqnarray*}
\widehat{f}_{\textrm{net}}\in \mcF(L, \bfp(T)),\quad \textrm{with $L=(2m+3)(k+d)+1$ and $T=2^{k+4}d(M+2k)^d$},
\end{eqnarray*}
where $k$ is the order of tensor product B-spline basis, $d$ is the dimension of explanatory variable $\bfX$ , $M= h^{-1}$ is the inverse of knots separation distance, $m$ is an integer characterizing the number of hidden layers of the network. All the results in this subsection are discussed  when $m, M, n$ diverge while assuming $k, d$ are fixed constant.

Theorem \ref{thm:asymptotic:normality:neural:main:text} below establishes a pointwise asymptotic distribution for
$\widehat{f}_{\textrm{net}}$.
\begin{theorem}\label{thm:asymptotic:normality:neural:main:text}
Under the Assumptions \ref{A0} and \ref{Assumption:A1}, if $k\geq \max(\beta, 2)$, $n^{\frac{1}{2\beta+d}}=o(M)$, $M^d\log(M)=o(n)$ and $nM^d=o(16^{m})$, then
 for any fixed point $\bfx \in \Omega$, we have
\begin{eqnarray*}
	\frac{\widehat{f}_{\textrm{net}}(\bfx)-f_0(\bfx)}{\sqrt{\bfD^T_k(\bfx)(\Phi^T\Phi)^{-1}\bfD_k(\bfx)}}\xrightarrow[]{D}N(0,1),
\end{eqnarray*}
where $\Phi=(\bfD_k(\bfX_1),\bfD_k(\bfX_2),\ldots, \bfD_k(\bfX_n))^T\in \mathbb{R}^{n\times q}$ with $q=(M+k-1)^d$.
\end{theorem}
\begin{proof}[{\it Proof of Theorem \ref{thm:asymptotic:normality:neural:main:text}}]
By (\ref{eq:counting:fnet}) and Assumption \ref{Assumption:A1}, we know $M= h^{-1}$ and
\begin{equation*}
\widehat{f}_{\textrm{net}}\in \mcF(L, \bfp(T)),\quad \textrm{with $L=(2m+3)(k+d)+1$ and $T=2^{k+4}d(M+2k)^d$}.
\end{equation*}
So the rate conditions are equivalent to $hn^{\frac{1}{2\beta+d}}=o(1)$, $\log(h^{-1})=o(nh^d)$ and $n^{1/2}h^{-d/2}=o(4^{m})$.

For fixed $\bfx\in\Omega$, let $V(\bfx)=\bfD^T_k(\bfx)(\Phi^T\Phi)^{-1}\bfD_k(\bfx)$.
By \cite[Theorems 3.1 and 5.2]{h03}, it follows that
\begin{eqnarray}\label{thm11:eqn:1}
	\frac{\widehat{f}_{\textrm{pilot}}(\bfx)-f_0(\bfx)}{\sqrt{V(\bfx)}}\xrightarrow[]{D}N(0,1).
\end{eqnarray}
It is well known  that the tensor product B-spline basis satisfies $\sum_{s=1}^qD_{\bfi_s, k}(\bfx)=1$ for all $\bfx \in \Omega$ (e.g., see Section 15 in \citealp{gkkw06}). Given a point $\bfx \in \Omega$, let us denote $\Gamma_\bfx=\{\bfi \in \Gamma| D_{\bfi,k}(\bfx)>0\}$. By the construction of $D_{\bfi,k}$, there are only $k^d$ basis functions among $D_{\bfi_1,k}(\bfx),\ldots,D_{\bfi_q,k}(\bfx)$ with positive values, while the rest are all zero. Hence, it follows that $|\Gamma_\bfx|=k^d$. The above fact implies that $\sum_{\bfi \in \Gamma_\bfx}D_{\bfi, k}(\bfx)=1$ and $\bfD_k^T(\bfx)\bfD_k(\bfx)=\sum_{\bfi \in \Gamma_\bfx}D^2_{\bfi, k}(\bfx)\geq |\Gamma_\bfx|^{-1}=k^{-d}$, where the equality holds when $D_{\bfi, k}(\bfx)=|\Gamma_\bfx|^{-1}$ for all $\bfi \in \Gamma_\bfx$. 

Lemma \ref{lemma:empirical:eigen:value} implies that with probability approaching 1, we have 
\begin{eqnarray*}
V(\bfx)&=&\bfD^T_k(\bfx)(\Phi^T\Phi)^{-1}\bfD_k(\bfx)\\
&\geq&\lambda_{\min}((\Phi^T\Phi)^{-1})\bfD_k(\bfx)^T\bfD_k(\bfx)\\
&=&\frac{1}{\lambda_{\max}(\Phi^T\Phi)}\bfD_k(\bfx)^T\bfD_k(\bfx)\\
&\geq& \frac{1}{2a_2nh^d}\bfD_k(\bfx)^T\bfD_k(\bfx)\geq  \frac{1}{2a_2k^{d}nh^d},
\end{eqnarray*}
where $a_2$ is the constant (\ref{eq:definition:Omegan}).
By Lemma \ref{thm:approximation:sieve:DNN} we get that
$|\widehat{f}_{\textrm{pilot}}(\bfx)-\widehat{f}_{\textrm{net}}(\bfx)|^2=O_P(h^{-2d}4^{-2m})$.
Therefore, 
\begin{equation}\label{thm11:eqn:2}
\frac{\widehat{f}_{\textrm{pilot}}(\bfx)-\widehat{f}_{\textrm{net}}(\bfx)}{\sqrt{V(\bfx)}}=O_P(n^{1/2}h^{-d/2}4^{-m})=o_P(1).
\end{equation}
Theorem \ref{thm:asymptotic:normality:neural:main:text} follows by (\ref{thm11:eqn:1}) and (\ref{thm11:eqn:2}). This completes the proof.
\end{proof}

In practice, it is often of interest to test whether $Y_i$ and $\bfX_i$ are statistically independent, equivalently, to test $f_0$ is constant. 
In what follows, we consider an elementary hypothesis testing problem: $H_0: f_0=0$ vs. $H_1: f\neq0$. In general, one can subtract the constant from $f_0$, or if the constant is unknown, subtract $\bar{Y}$ from $f_0$, and test the difference equals zero.
Consider a test statistic $T_n=\|\widehat{f}_{\textrm{net}}\|_n^2$, where $\|f\|_n^2=\sum_{i=1}^n f(\bfx_i)^2/n$
is the empirical norm. It should be mentioned that $T_n$ relies on $m, M$ since $\widehat{f}_{\textrm{net}}$ does.
The following Theorem \ref{thm:optimal:test:neural:main:text} is a byproduct of Lemma \ref{thm:approximation:sieve:DNN},
which derives null distribution of $T_n$ and analyzes its power under a sequence of local alternatives.
%
\begin{theorem}\label{thm:optimal:test:neural:main:text}
Under the Assumptions \ref{A0} and \ref{Assumption:A1}, if $k\geq \max(\beta,2)$, $n^2M^d=O(16^m)$ and $M\asymp n^{\frac{2}{4\beta+d}}$, then the following hold:
\begin{enumerate}
\item Under $H_0: f_0=0$, it follows that 
\begin{eqnarray}\label{thm:11:1}
	\frac{nT_n-q}{\sqrt{2q}}\xrightarrow[]{D}N(0,1),
\end{eqnarray}
where $q=(M+k-1)^d$.
\item For any $\delta>0$, there exists a $C_\delta>0$ such that, under $H_1: f=f_0$ with 
$\|f_0\|_n\geq C_\delta n^{-\frac{2\beta}{4\beta+d}}$, it holds that
\begin{eqnarray}\label{thm:11:2}
	\pr\bigg(\bigg|\frac{nT_n-q}{\sqrt{2q}}\bigg|>z_{\alpha/2}\bigg)\geq 1-\delta,
\end{eqnarray}
where $z_{\alpha/2}$ is the $1-\alpha/2$ upper percentile of standard normal variable.
\end{enumerate}
\end{theorem}

Part (\ref{thm:11:1}) of Theorem \ref{thm:optimal:test:neural:main:text} suggests a testing rule at significance $\alpha$:
reject $H_0$ if and only if 
\[
\bigg|\frac{nT_n-q}{\sqrt{2q}}\bigg|\ge z_{\alpha/2}.
\]
Part (\ref{thm:11:2}) of Theorem \ref{thm:optimal:test:neural:main:text}
says that the power of $T_n$ is at least $1-\delta$ provided that the
null and alternative hypotheses are separated by $C_\delta n^{-\frac{2\beta}{4\beta+d}}$
in terms of $\|\cdot\|_n$-norm.
The separation rate is optimal in the sense of \cite{Ingster93}.
\begin{proof}[{\it Proof of Theorem \ref{thm:optimal:test:neural:main:text}}]
The proof consists of two steps. The first step is to establish the asymptotic distribution of the test statistic based $\widehat{f}_{\textrm{pilot}}$, while the second step is to show that the test statistic $T_n$ has the same limiting distribution. By (\ref{eq:counting:fnet}) and Assumption \ref{Assumption:A1}, we know $M= h^{-1} and$
\begin{equation*}
\widehat{f}_{\textrm{net}}\in \mcF(L, \bfp(T)),\quad \textrm{with $L=(2m+3)(k+d)+1$ and $T=2^{k+4}d(M+2k)^d$}.
\end{equation*}
So the rate conditions are equivalent to $nh^{-d/2}4^{-m}=o(1)$ and $h\asymp n^{-\frac{2}{4\beta+d}}$.

Step 1: Using the notation in the proof of Lemma \ref{thm:approximation:sieve:DNN} and by (\ref{eq:lemma:rate:of:convergence:sieve:eq0}), we have
\begin{eqnarray*}
	\widehat{\bff}_{\textrm{pilot}}&=&\Phi(\Phi^T \Phi)^{-1}\Phi^T \bff_0+\Phi(\Phi^T \Phi)^{-1}\Phi^T \bfepsilon.
\end{eqnarray*}
Under $H_0: f_0=0$, it follows that $\widehat{\bff}_{\textrm{pilot}}^T\widehat{\bff}_{\textrm{pilot}}=\bfepsilon^T\Phi(\Phi^T \Phi)^{-1}\Phi^T \bfepsilon$ and
\begin{eqnarray*}
	\widehat{\bff}_{\textrm{pilot}}^T\widehat{\bff}_{\textrm{pilot}} | \mathbb{X} \sim \chi^2(q),
\end{eqnarray*}
where we used the fact that $\epsilon_i$ are i.i.d normal and is free of $\mathbb{X}$.
Since $q=(M+k-1)^d \asymp h^{-d} \to \infty$, we conclude from central limit theorem that
\begin{eqnarray}
	\frac{\widehat{\bff}_{\textrm{pilot}}^T\widehat{\bff}_{\textrm{pilot}}-q}{\sqrt{2q}}\xrightarrow[]{D}N(0,1).\label{eq:thm:optimal:test:neural:main:text:eq1}
\end{eqnarray}

Suppose that $f_0$ satisfies $\|f_0\|_n\geq C_\delta\gamma_n$ with $\gamma_n=n^{-\frac{2\beta}{4\beta+d}}$ for some $C_\delta$ large enough. 
Then it follows that
\begin{eqnarray*}
\widehat{\bff}_{\textrm{pilot}}^T\widehat{\bff}_{\textrm{pilot}}=\bff_0^T\Phi(\Phi^T \Phi)^{-1}\Phi^T \bff_0+2\bff_0^T\Phi(\Phi^T \Phi)^{-1}\Phi^T \bfepsilon+\bfepsilon^T\Phi(\Phi^T \Phi)^{-1}\Phi^T \bfepsilon\equiv S_1+2S_2+S_3.
\end{eqnarray*}
By simple algebra, we show that
\begin{eqnarray*}
	\bff_0^T(I-\Phi(\Phi^T \Phi)^{-1}\Phi^T)\bff_0&=&(\Phi C+\bfE)^T(I-\Phi(\Phi^T \Phi)^{-1}\Phi^T)(\Phi C+\bfE)\\
	&=&\bfE^T(I-\Phi(\Phi^T \Phi)^{-1}\Phi^T)\bfE\\
	&\leq& \bfE^T\bfE\leq A_{f_0}^2nh^{2\beta}.
\end{eqnarray*}
As a consequence it follows that
\begin{eqnarray*}
	S_1=\bff_0^T\bff_0-\bff_0^T(I-\Phi(\Phi^T \Phi)^{-1}\Phi^T)\bff_0
	\geq C_\delta^2n\gamma_n^2-A_{f_0}^2nh^{2\beta}	=C_\delta^2 n^{\frac{d}{4\beta+d}}-A_{f_0}^2nh^{2\beta}.
\end{eqnarray*}
Since $h=M^{-1} \asymp n^{-\frac{2}{4\beta+d}}$, it follows that $nh^{2\beta}\asymp n^{\frac{d}{4\beta+d}}$. If we choose $C_\delta>0$ large enough, it implies that $S_1= \frac{1}{2}C_\delta^2n^{\frac{d}{4\beta+d}}$, which leads to 
\begin{eqnarray*}
	\frac{S_1}{\sqrt{2q}}\geq \frac{1}{2\sqrt{2q}}C_\delta^2 n^{\frac{d}{4\beta+d}} \asymp \frac{1}{2\sqrt{2}}C_\delta^2n^{\frac{d}{4\beta+d}}h^{\frac{d}{2}}\asymp \frac{1}{2\sqrt{2}}C_\delta^2 \quad \textrm{ and }\quad \sqrt{\frac{S_1}{2q}}\to 0.
\end{eqnarray*}
Here the condition $q=(M+k-1)^d\asymp h^{-d}$ is used.
So $\sqrt{\frac{S_1}{2q}}\leq \frac{1}{4C_\delta}\frac{S_1}{\sqrt{2q}}$ for $n$ large enough.
Taking conditional expectation, we have 
\begin{eqnarray*}
	\pr\left(|S_2|^2>C_\delta^2S_1|\mathbb{X}\right)=P(|Z|>C_\delta)\leq \delta,
\end{eqnarray*}
where $Z$ is standard normal random variable and the last inequality holds with large $C_\delta$.
Therefore, we have that 
\begin{eqnarray}
	&&\pr\bigg(\bigg|\frac{\widehat{\bff}_{\textrm{pilot}}^T\widehat{\bff}_{\textrm{pilot}}-q}{\sqrt{2q}}\bigg|\leq Z_{\alpha/2}\bigg)\nonumber\\	
	&=&\pr\bigg(\bigg|\frac{S_3-q}{\sqrt{2q}}+\frac{S_1}{\sqrt{2q}}+\frac{2S_2}{2q}\bigg|\leq Z_{\alpha/2}\bigg)\nonumber\\
	&\leq& \pr\bigg(\bigg|\frac{S_3-q}{\sqrt{2q}}+\frac{S_1}{\sqrt{2q}}+\frac{2S_2}{\sqrt{2q}}\bigg|\leq Z_{\alpha/2}, |S_2|\leq C_\delta \sqrt{S_1}\bigg)+P\bigg(|S_2|>C_\delta\sqrt{S_1}\bigg). \label{eq:lemma:optimal:test:sieve:eq1}
\end{eqnarray}
By the choice of $C_\delta$, the second term in (\ref{eq:lemma:optimal:test:sieve:eq1}) is bounded by $\delta$, while the first term yields following inequality:
\begin{eqnarray*}
	&&\pr\bigg(\bigg|\frac{S_3-q}{\sqrt{2q}}+\frac{S_1}{\sqrt{2q}}+\frac{2S_2}{\sqrt{2q}}\bigg|\leq Z_{\alpha/2}, |S_2|\leq C_\delta \sqrt{S_1}\bigg)\\
	&=&\pr\bigg(-Z_{\alpha/2}- \frac{S_1}{\sqrt{2q}}-\frac{2S_2}{\sqrt{2q}}\leq \frac{S_3-q}{\sqrt{2q}}\leq Z_{\alpha/2}- \frac{S_1}{\sqrt{2q}}-\frac{2S_2}{\sqrt{2q}} , |S_2|\leq C_\delta \sqrt{S_1}\bigg)\\
	&\leq&\pr\bigg(-Z_{\alpha/2}- \frac{S_1}{\sqrt{2q}}-\frac{2C_\delta \sqrt{S_1}}{\sqrt{2q}}\leq \frac{S_3-q}{\sqrt{2q}}\leq Z_{\alpha/2}- \frac{S_1}{\sqrt{2q}}+\frac{2C_\delta \sqrt{S_1}}{\sqrt{2q}} , |S_2|\leq C_\delta \sqrt{S_1}\bigg)\\
&\leq&\pr\bigg(-Z_{\alpha/2}- \frac{3S_1}{2\sqrt{2q}}\leq \frac{S_3-q}{\sqrt{2q}}\leq Z_{\alpha/2}- \frac{S_1}{2\sqrt{2q}}\bigg)\\
&\leq&\pr\bigg(\frac{S_3-q}{\sqrt{2q}}\leq Z_{\alpha/2}-\frac{C_\delta^2}{2\sqrt{2}}\bigg).
\end{eqnarray*}
Combining above and taking limit on both sides, it follows that
\begin{eqnarray}
	\lim_{n \to \infty}\pr\bigg(\bigg|\frac{\widehat{\bff}_{\textrm{pilot}}^T\widehat{\bff}_{\textrm{pilot}}-q}{\sqrt{2q}}\bigg|\leq Z_{\alpha/2}\bigg)\leq \pr\bigg(Z\leq Z_{\alpha/2}-\frac{C_\delta^2}{2\sqrt{2}}\bigg)\leq \delta. \label{eq:thm:optimal:test:neural:main:text:eq2}
\end{eqnarray}

Step 2:
Observe that
\begin{eqnarray}\label{eq:them:optimal:test:neural}
	\frac{n\|\widehat{f}_{\textrm{net}}\|_n^2-q}{\sqrt{2q}}=\frac{n\|\widehat{f}_{\textrm{pilot}}\|_n^2-q}{\sqrt{2q}}+\frac{n\|\widehat{f}_{\textrm{net}}\|_n^2-n\|\widehat{f}_{\textrm{pilot}}\|_n^2}{\sqrt{2q}}.
\end{eqnarray}
By Lemma \ref{lemma:rate:of:convergence:sieve} and Lemma \ref{thm:approximation:sieve:DNN}, 
both $\|\widehat{f}_{\textrm{net}}-\widehat{f}_{\textrm{pilot}}\|_n$ and $\|\widehat{f}_{\textrm{pilot}}-f_0\|_n$
are $O_P(1)$, and we have
\begin{eqnarray*}
	|\|\widehat{f}_{\textrm{net}}\|_n^2-\|\widehat{f}_{\textrm{pilot}}\|_n^2|&=&
	|\|\widehat{f}_{\textrm{net}}\|_n-\|\widehat{f}_{\textrm{pilot}}\|_n|\times \left(\|\widehat{f}_{\textrm{net}}\|_n+\|\widehat{f}_{\textrm{pilot}}\|_n\right)\\
	&\leq&\|\widehat{f}_{\textrm{net}}-\widehat{f}_{\textrm{pilot}}\|_n\times \left(\|\widehat{f}_{\textrm{net}}-\widehat{f}_{\textrm{pilot}}\|_n+2\|\widehat{f}_{\textrm{pilot}}\|_n\right)\\
	&\leq&\|\widehat{f}_{\textrm{net}}-\widehat{f}_{\textrm{pilot}}\|_n\times 
	\left(\|\widehat{f}_{\textrm{net}}-\widehat{f}_{\textrm{pilot}}\|_n+2\|\widehat{f}_{\textrm{pilot}}-f_0\|_n+2\|f_0\|_n\right)\\
	&=&\|\widehat{f}_{\textrm{net}}-\widehat{f}_{\textrm{pilot}}\|_n\times O_P(1)\\
	&=&O_P(h^{-d}4^{-m}).
\end{eqnarray*}
Therefore, the second term in (\ref{eq:them:optimal:test:neural})
is of order $O_P(nh^{-d}4^{-m}q^{-1/2})=O_P(nh^{-d/2}4^{-m})=o_P(1)$, 
where we have used the fact $q=(M+k-1)^d\asymp h^{-d}$.
The result then follows by (\ref{eq:thm:optimal:test:neural:main:text:eq1}) and (\ref{eq:thm:optimal:test:neural:main:text:eq2}) . This completes the proof.
\end{proof}

\section{Network Approximation to Additive Model}\label{sec:additive:model}
The optimal rate in Theorem \ref{thm:neural:estimator:rate:of:convergence:main:text} suffers from the `curse' of dimensionality.  In this section, we show that  this issue can be addressed when $f_0$ has an additive structure. Specifically, let us consider the following function space:
\begin{eqnarray}
	\Lambda^{\boldsymbol\beta}_+(F, \Omega)=\left\{f: \Omega \to \mathbb{R}|\; f(\bfx)=a+\sum_{j=1}^dg_{j}(x_j) \textrm{ with } g_j \in \Lambda^{\beta_j}(F, [0,1]) \textrm{ and }\int_0^1 g_j(x)dx=0 \right\}, \nonumber
\end{eqnarray}
where $F>0$ is the radius, and $\boldsymbol\beta=(\beta_1, \ldots, \beta_d)\in (0,\infty)^d$ are the degrees of smoothness for $g_j$'s. Clearly, any $f\in\Lambda^{\bm{\beta}}_+(F, \Omega)$ has an expression $f(\bfx)=a+\sum_{j=1}^dg_{j}(x_j)$
with the $j$th additive component belonging to the ball of univariate $\beta_j$-H\"{o}lder functions with radius $F$. Moreover, the constraint  $\int_0^1 g_{j,0}(x)dx=0$  is  to avoid identifiability issue.

\begin{theorem}\label{thm:rate:convergence:additive:main:text}
Let Assumption \ref{A0} be satisfied. Suppose that $L\to \infty$, $T\to\infty$ and $T\log T=o(n)$ as $n\to\infty$, then for any fixed constant $F>0$ and vector $\boldsymbol\beta=(\beta_1, \ldots, \beta_d)\in (0,\infty)^d$, it follows that
\begin{eqnarray*}
	\inf_{\widehat{f}\in\mathcal{F}(L,\bfp(T))}\sup_{f_0\in \Lambda^{\boldsymbol\beta}_+(F, \Omega)}
	\ev_{f_0}\bigg(\|\widehat{f}-f_0\|_{L^2}^2\bigg|\mathbb{X}\bigg)=O_P\bigg(\frac{1}{T^{2\beta^*}}+\frac{T}{n}+\frac{T^2}{2^{\frac{L}{1+k}}}\bigg),
\end{eqnarray*}
where $\beta^*=\min_{1\le j\le d}\beta_j$,
$k$ is the smallest integer satisfying $k\geq \max(\beta_1,\ldots,\beta_d, 2)$, and the $O_P$ is in the sense that $T, L, n$ are diverging.
Hence, if $T\asymp n^{\frac{1}{2\beta^*+1}}$  and $n^{\frac{2\beta^*+2}{2\beta^*+1}}=O(2^{\frac{L}{1+k}})$, then
\begin{eqnarray*}
\inf_{\widehat{f}\in\mathcal{F}(L,\bfp(T))}\sup_{f_0\in \Lambda^{\boldsymbol\beta}_+(F, \Omega)}\ev_{f_0}\bigg(\|\widehat{f}-f_0\|_{L^2}^2\bigg|\mathbb{X}\bigg)=O_P\left(n^{-\frac{2\beta^*}{2\beta^*+1}}\right).
\end{eqnarray*}
\end{theorem}
The rate $n^{-\frac{2\beta^*}{2\beta^*+1}}$ in Theorem \ref{thm:rate:convergence:additive:main:text} is optimal 
in nonparmetric additive estimation. 
When $\beta_1=\cdots=\beta_d=\beta$, the rate simply becomes $n^{-\frac{2\beta}{2\beta+1}}$
whose optimality has been proved by \cite{s85}.
Otherwise, the optimal rate relies on the least order of smoothness of the $d$ univariate functions. 

The rest part of this section is devoted to proving Theorem \ref{thm:rate:convergence:additive:main:text}.
Throughout we keep in mind that the true regression function $f_0$ admits an additive expression
\begin{eqnarray*}
f_0(\bfx)=f_0(x_1,\ldots,x_d)=\alpha_0+g_{1,0}(x_1)+\ldots+g_{d,0}(x_d),
\end{eqnarray*}
where $\alpha_0$ is an unknown constant.  Before proving the theorem, let us settle down some notation.
For $j=1, 2,\ldots, d$, given integers $M_j, k_j\geq 2$ and knots $t_{-k_j+1, j}<t_{-k_j+2, j}< \ldots <t_{0,j}<t_{1,j}<\ldots<t_{M_j,j}<t_{M_j+1,j}<\ldots<t_{M_j+k_j+1,j} $ with $t_{0,j}=0, t_{M_j,0}=1$, let $\bfB_{k_j,j}(x)\in \mathbb{R}^{M_j+k_j-1}$ denote the vector of 
univariate B-spline basis functions (with respect to variable $x_j$). 
Since the collection of these univariate B-spline basis does not form a basis on the additive function space
due to the sum-to-one condition, we instead use the following polynomial spline basis to approximate the additive components $g_{j,0}$'s:
\begin{eqnarray*}
	\bfP_{k_j,j}(x)=\bigg(x, x^2, \ldots, x^{k_j-1}, (x-t_{1,j})_+^{k_j-1}, \ldots, (x-t_{M_j-1,j})_+^{k_j-1}\bigg)^T\in \mathbb{R}^{M_j+k_j-2},
	j=1,\ldots,d.
\end{eqnarray*}
The central idea is the approximation 
$f_0(x_1,\ldots,x_d)\approx a+\sum_{j=1}^d W_j^T\bfP_{k_j,j}(x_j)$
for some constants $a\in\mathbb{R}$ and $W_j\in\mathbb{R}^{M_j+k_j-2}$. 
By least square estimation, an estimator of $f_0$ is 
\begin{eqnarray*}
\widehat{f}_{\textrm{pilot}}(x_1,\ldots,x_d)
=\widehat{a}+\sum_{j=1}^d\widehat{f}_j(x_j) \textrm{ with } \widehat{f}_j(x)=\widehat{W}_j^T\bfP_{k_j,j}(x).
\end{eqnarray*}
If we define the centralized estimator 
$\widehat{g}_j(x)=\widehat{f}_j(x)-\int_0^1\widehat{f}_j(u)du$, then it turns out to be a consistent estimator of $g_{j,0}$; see Lemma \ref{lemma:rate:convergece:individual:additive}, and we have
\begin{eqnarray}
\widehat{f}_{\textrm{pilot}}(x_1,\ldots,x_d)
=\widehat{\alpha}+\sum_{j=1}^d\widehat{g}_j(x_j) \textrm{ with } \widehat{\alpha}=\widehat{a}+\sum_{j=1}^d\int_0^1\widehat{f}_j(u)du. \label{eq:additive:alphahat}
\end{eqnarray}
Note that $\bfB_{k_j, j}$ is the B-spline basis. So $\widehat{g}_j$ can be written as $\widehat{C}_j^T\bfB_{k_j,j}(x)$ for some $\widehat{C}_j\in \mathbb{R}^{M_j+k_j-1}$, we define a neural network estimator $\widetilde{g}_j(x)=\widehat{C}_j^T\widetilde{\bfB}_{k_j,j}(x)$ for $j=1,\ldots, d$ and 
\begin{eqnarray}\label{eq:expression:fnet:additive}
\widehat{f}_{\textrm{net}}(\bfx)=\widehat{\alpha}+\sum_{j=1}^d\widetilde{g}_j(x_j).
\end{eqnarray}
By similar argument as (\ref{eq:counting:fnet}),  for any integers $k_j, M_j, m\geq 2$, we can construct the network satisfying
\begin{equation}\label{eq:counting:fnet:additive}
\widehat{f}_{\textrm{net}}\in \mcF(L, \bfp(T))\quad \textrm{with $L=(2m+3)\max_{1\leq j\leq d}(k_j+1)+1$ and $T=\sum_{j=1}^d2^{k_j+4}(M_j+2k_j)$}.
\end{equation}
Moreover, the following notation plays a similar role as that in the proof of Theorem \ref{thm:neural:estimator:rate:of:convergence:main:text}:
\begin{eqnarray}
q_+&=&1+\sum_{j=1}^d(M_j+k_j-2),\nonumber\\
\bfP(\bfx)&=&(1, \bfP^T_{k_1,1}(x_1),\bfP^T_{k_2,2}(x_2),\ldots, \bfP^T_{k_d,d}(x_d))^T\in \mathbb{R}^{q_+},\nonumber\\
\Phi_+&=&(\bfP(\bfX_1), \bfP(\bfX_2),\ldots, \bfP(\bfX_n))^T\in \mathbb{R}^{n\times q_+},\nonumber\\
	\Theta_n^+&=&\{f(\bfx)| f(\bfx)=a+\sum_{j=1}^d g_j(x_j) \textrm{ with } a\in \mathbb{R}, g_j(x)=b_j^T\bfP_{k_j,j}(x),\nonumber\\
	&& \int_0^1 g_j(x)dx=0 \textrm{ for some } b_j\in \mathbb{R}^{M_j+k_j-2} \textrm{ and } j=1,\ldots, d\},\nonumber\\
	\Omega_n^+&=&\{\|g\|_{L^2}^2/2\leq \|g\|_n^2\leq 2\|g\|_{L^2}^2, \textrm{ for all } g\in \Theta_n^+\}.\label{eq:symbol:additive}
\end{eqnarray}  
To handle the additive model, we introduce a new norm of a function $g$ as $\|g\|^2=\int_{\Omega}g^2(\bfx)dx$. We would like to comment that another norm used in previous sections is  $\|g\|^2_{L^2}=\int_{\Omega}g^2(\bfx)Q(\bfx)dx$, which are equivalent to $\|\cdot\|$ under Assumption \ref{A0}.
Finally, we will need the following assumption during the proof, which is in the similar spirit of Assumption \ref{Assumption:A1}.
\begin{Assumption}\label{Assumption:A2}
For $j=1,\ldots,d$, the order of B-spline satisfies $k_j\geq \beta_j$, and the knots $\{t_{i,j}, i=-k_j+1, \ldots, M_j+k_j+1\}$ are equally separated by
constant $h_j=M_j^{-1}$. In the analysis, we need $M_j\to \infty$ and $h_j\to 0$ for all $j=1\,\ldots, d$.
\end{Assumption}
\begin{proposition}\label{proposition:constant:plus:zero:mean:function:An}
Suppose that $g_0$ is a constant function and $g_1$ is a measurable function satisfying $\int_\Omega g_1(\bfx)d\bfx=0$. 
Moreover, $\|g_1\|_{\textrm{sup}} \leq K\|g_1\|$ for some constant $K>0$. Then
$\|g_0+g_1\|_{\textrm{sup}}\leq (K+2)\|g_0+g_1\|$.
\end{proposition}
\begin{proof}[{\it Proof of Proposition \ref{proposition:constant:plus:zero:mean:function:An}}]
Observe that for any constant function $g_0$, we have
$\|g_1\|=\|g_1+g_0\|=\|g_1\|^2+g_0^2$. Moreover, Assumption \ref{A0} leads to that, 
for some $c>1$ and all $g$ with $\|g\|_{L^2}<\infty$,  
it holds that $c^{-1}\|g\|^2\leq\|g\|_{L^2}^2\leq c\|g\|^2$. Therefore, we have
\begin{eqnarray*}
	\|g_0+g_1\|_{\textrm{sup}} &\leq& \|g_0\|_{\textrm{sup}}+\|g_1\|_{\textrm{sup}}\\
	&\leq&\|g_0\|+K\|g_1\|\\
	&\leq& \|g_0+g_1\|+\|g_1\|+K\|g_1+g_0\|\\
	&\leq& \|g_0+g_1\|+\|g_1+g_0\|+K\|g_1+g_0\|\\
	&\leq& (K+2)\|g_0+g_1\|.
\end{eqnarray*}
Proof is complete.
\end{proof}

\begin{lemma}\label{lemma:equivalence:norm:additive:model}
Suppose Assumptions \ref{A0} and \ref{Assumption:A2} hold with integers $k_j\geq \max(\beta_j, 2)$. Moreover, if the sequences in Assumption \ref{Assumption:A2} satisfies  $nh_j^2 \to \infty$ and $h_j \to 0$ for each $j=1,2,\ldots, d$, then  the following holds
\begin{eqnarray*}
	\sup_{g\in \Theta_n^+}\bigg|\frac{\|g\|_n^2}{\|g\|_{L^2}^2}-1\bigg|=o_P(1),
\end{eqnarray*}
where $\Theta_n^+$ is the function space defined in (\ref{eq:definition:Omegan}).
As a consequence, it follows that $\pr(\Omega_n^+)\to 1$. Here $\Omega_n^+$ is the event defined in (\ref{eq:definition:Omegan}).
\end{lemma}
\begin{proof}[{\it Proof of Lemma \ref{lemma:equivalence:norm:additive:model}}]
Let $g(\bfx)=\sum_{j=1}^dg_j(x_j)$, where $g_j$ satisfies $\int_0^1 g_j(x)dx=0$ for $j=1,\ldots, d$.
By \cite[Theorem 5.1.2]{dl93} we get that $\|g_j\|_{\textrm{sup}}\leq A_j \|g_j\|$ with $A_j \asymp h_j^{-1/2}$. Direct examination shows that
\begin{eqnarray*}
	\|g\|_{\textrm{sup}}\leq \sum_{j=1}^d\|g_j\|_{\textrm{sup}} \leq \sum_{j=1}^d A_j\|g_j\|\leq 
	\left(\sum_{j=1}^dA_j^2\right)^{1/2}\left(\sum_{j=1}^d \|g_{j}\|^2\right)^{1/2}\leq 
	\left(\sum_{j=1}^dA_j^2\right)^{1/2}\left(c_d\|g\|^2\right)^{1/2},
\end{eqnarray*}
where the last inequality follows from Lemma 3.6 of \cite{s94} and $c_d$ is a constant depending on $d$ only. 
Applying Proposition \ref{proposition:constant:plus:zero:mean:function:An} and by Assumption \ref{A0}, we obtain that
\begin{eqnarray*}
	\|f\|_{\textrm{sup}}\leq \bigg(\bigg(c_d\sum_{j=1}^dA_j^2\bigg)^{1/2}+2\bigg)\|f\|_{2}\leq c\bigg(\bigg(c_d\sum_{j=1}^dA_j^2\bigg)^{1/2}+2\bigg)\|f\|_{L^2}^2, \textrm{ for all } f\in \Theta_n^+.
\end{eqnarray*}
The dimension of $\Theta_n^+$, $q_{+}\leq \sum_{j=1}^d(M_j+k_j-1)+1\asymp \sum_{j=1}^d h_j^{-1}$. Therefore, by Lemma 2.3 in \cite{h03} and rate conditions given, we prove the result. 
\end{proof}

\begin{lemma}\label{lemma:invertible:empirical:design:matrix:additive}
Suppose Assumptions \ref{A0} and \ref{Assumption:A2} hold with integers $k_j\geq \max(\beta_j, 2)$. Moreover, if the sequences in Assumption \ref{Assumption:A2} satisfies  $nh_j^2 \to \infty$ and $h_j \to 0$ for each $j=1,2,\ldots, d$, then on event $\Omega_n^+$, 
$\Phi_+^T\Phi_+$ is invertible.
\end{lemma}
\begin{proof}[{\it Proof of Lemma \ref{lemma:invertible:empirical:design:matrix:additive}.}]
Let $\widehat{\bfB}=n^{-1}\Phi_+^T\Phi_+$ and $\bfB=\int \bfP(\bfx)\bfP(\bfx)^TQ(\bfx)d\bfx$. For $g(\bfx)=u^T\bfP(\bfx)$, we have
$u^T\widehat{\bfB}u=\|g\|_n^2$ and $u^T\bfB u=\|g\|^2$.
On event $\Omega_n^+$, since $\bfB$ is positive definite, $\widehat{\bfB}$ is also positive definite. Proof is complete.
\end{proof}

\begin{lemma}\label{lemma:convergence:addtive:sieve}
Suppose Assumptions \ref{A0} and \ref{Assumption:A2} hold with integers $k_j\geq \max(\beta_j, 2)$. Moreover, if the sequences in Assumption \ref{Assumption:A2} satisfies  $nh_j^2 \to \infty$ and $h_j \to 0$ for each $j=1,2,\ldots, d$, then the following holds uniformly for all $f_0\in \Lambda^{\bm{\beta}}_+(F, \Omega)$ on event $\Omega_n^+$:
\begin{eqnarray*}
	\ev_{f_0}\bigg( \|\widehat{f}_{\textrm{pilot}}-f_0\|_{L^2}^2\bigg|\mathbb{X}\bigg)\leq  2^{d+3} \sum_{j=1}^d A_{g_{j,0}}^2h_j^{2\beta_j}+\frac{8q_+}{n}.
\end{eqnarray*}
\end{lemma}
\begin{proof}[{\it Proof of Lemma \ref{lemma:convergence:addtive:sieve}}]
For any $f_0\in\Lambda^{\bm{\beta}}_+(F, \Omega)$, let $\mathbf{f}_0=(f_0(\bfX_1),\ldots,f_0(\bfX_n))^T$,
$\bm{\epsilon}=(\epsilon_1,\ldots,\epsilon_n)^T$, and
$\widehat{\bff}_{\textrm{pilot}}=(\widehat{f}_{\textrm{pilot}}(\bfX_1),
\ldots,\widehat{f}_{\textrm{pilot}}(\bfX_n))^T$. 
According to Lemma \ref{lemma:spline:approximation} and the condition $k_j\geq \max(\beta_j,2)$ for $j=1,\ldots, d$, 
there exists a vector $W \in \mathbb{R}^{q_+}$ 
such that
$\sup_{\bfx \in \Omega}|W^T\bfP(\bfx)-f_0(\bfx)|\leq \sum_{j=1}^dA_{g_{j,0}}h_j^{\beta_j}$, where the constant $A_{g_{j,0}}$ satisfies $\sup_{g_{j,0}\in \Lambda^{\beta_j}(F, [0, 1])}A_{g_{j,0}}<\infty$. For simplicity, we further define $f^*(\bfx)=W^T\bfP(\bfx)$ and $\bff^*=(f^*(\bfX_1),\ldots, f^*(\bfX_n))^\top$.

By Lemma \ref{lemma:invertible:empirical:design:matrix:additive} and similar argument in (\ref{eq:lemma:rate:of:convergence:sieve:eq0}), it follows  on event $\Omega_n^+$ that
\begin{eqnarray*}
\widehat{\bff}_{\textrm{pilot}}&=&\Phi_+(\Phi_+^T\Phi_+)^{-1}\Phi_+^T\bfY\\
&=&\Phi_+(\Phi_+^T\Phi_+)^{-1}\Phi_+^T(\Phi_+W+\bfE+\bfepsilon)\\
&=&\Phi_+W+\Phi_+(\Phi_+^T\Phi_+)^{-1}\Phi_+^T\bfE+\Phi_+(\Phi_+^T\Phi_+)^{-1}\Phi_+^T\bfepsilon\\
&=&\bff^*+\Phi_+(\Phi_+^T\Phi_+)^{-1}\Phi_+^T\bfE+\Phi_+(\Phi_+^T\Phi_+)^{-1}\Phi_+^T\bfepsilon,
\end{eqnarray*}
where $\bfE=\bff_0-\bff^*$. As a consequence, we have
\begin{eqnarray*}
\|\widehat{f}_{\textrm{pilot}}-f^*\|_n^2&=&\frac{1}{n}(\widehat{\bff}_{\textrm{pilot}}-\bff^*)^T(\widehat{\bff}_{\textrm{pilot}}-\bff^*)\\
&\leq& \frac{2}{n}\bfE^T \Phi_+(\Phi_+^T\Phi_+)^{-1}\Phi_+^T\bfE+\frac{2}{n}\bfepsilon^T\Phi_+(\Phi_+^T\Phi_+)^{-1}\Phi_+^T\bfepsilon\\
&\leq& \frac{2}{n}\bfE^T\bfE+\frac{2}{n}\bfepsilon^T\Phi_+(\Phi_+^T\Phi_+)^{-1}\Phi_+^T\bfepsilon\\
&\leq&2(\sum_{j=1}^dA_{g_{j,0}}h_j^{\beta_j})^2+\frac{2}{n}\bfepsilon^T\Phi_+(\Phi_+^T\Phi_+)^{-1}\Phi_+^T\bfepsilon\\
&\leq&2^d \sum_{j=1}^d A_{g_{j,0}}^2h_j^{2\beta_j}+\frac{2}{n}\bfepsilon^T\Phi_+(\Phi_+^T\Phi_+)^{-1}\Phi_+^T\bfepsilon,
\end{eqnarray*}
where we use the fact that $\Phi_+^T\Phi_+$ is invertible on $\Omega_n^+$ by Lemma \ref{lemma:invertible:empirical:design:matrix:additive}. By independence of $\bfepsilon$ and $\Phi_+$, it follows that on event $\Omega_n^+$,
\begin{eqnarray*}
	\ev_{f_0}\bigg(\bfepsilon^T\Phi_+(\Phi_+^T\Phi_+)^{-1}\Phi_+^T\bfepsilon\bigg| \mathbb{X}\bigg)=\textrm{Tr}\left(\Phi_+(\Phi_+^T\Phi_+)^{-1}\Phi_+^T\right)=q_{+}.
\end{eqnarray*}
Combining the above two inequalities and using the definition of $\Omega_N^+$, we show that
\begin{eqnarray*}
\ev_{f_0}\bigg( \|\widehat{f}_{\textrm{pilot}}-f^*\|_{L^2}^2\bigg|\mathbb{X}\bigg)\leq2\ev_{f_0}\bigg( \|\widehat{f}_{\textrm{pilot}}-f^*\|_n^2\bigg|\mathbb{X}\bigg)
\leq 2^{d+1} \sum_{j=1}^d A_{g_{j,0}}^2h_j^{2\beta_j}+\frac{4q_+}{n},
\end{eqnarray*}
which further implies that
\begin{eqnarray*}
	\ev_{f_0}\bigg( \|\widehat{f}_{\textrm{pilot}}-f_0\|_{L^2}^2\bigg|\mathbb{X}\bigg)&\leq& 2\ev_{f_0}\bigg( \|\widehat{f}_{\textrm{pilot}}-f^*\|_{L^2}^2\bigg|\mathbb{X}\bigg)+2\ev_{f_0}\bigg( \|f_0-f^*\|_{L^2}^2\bigg|\mathbb{X}\bigg)\\
	&\leq& 2^{d+2} \sum_{j=1}^d A_{g_{j,0}}^2h_j^{2\beta_j}+\frac{8q_+}{n}+2^{d+1} \sum_{j=1}^d A_{g_{j,0}}^2h_j^{2\beta_j}\\
	&\leq&2^{d+3} \sum_{j=1}^d A_{g_{j,0}}^2h_j^{2\beta_j}+\frac{8q_+}{n}.
\end{eqnarray*}
Proof is complete.
\end{proof}

\begin{proposition}\label{proposition:upper:bound:sum:norm}
Under Assumption \ref{A0}, if $g(\bfx)=a+\sum_{j=1}^d g_{j}(x_j)$ with $\int_0^1g_j(x)dx=0$, then it follows that
$\|g\|_{L^2}^2\geq a_3^d(a^2+\sum_{j=1}^d\|g_j\|_{L^2}^2)$,
where the constant $a_3>0$ only relies on the density $Q$.
\end{proposition}
\begin{proof}[Proof of Proposition \ref{proposition:upper:bound:sum:norm}]
This is a direct consequence of Lemma 3.1 in \cite{s94} and Assumption \ref{A0}.
\end{proof}

\begin{lemma}\label{lemma:rate:convergece:individual:additive}
Suppose Assumptions \ref{A0} and \ref{Assumption:A2} hold with integers $k_j\geq \max(\beta_j, 2)$. Moreover, if the sequences in Assumption \ref{Assumption:A2} satisfies  $nh_j^2 \to \infty$ and $h_j \to 0$ for each $j=1,2,\ldots, d$, then the following statement hold uniformly for all $f_0\in \Lambda^{\beta}_+(F, \Omega)$ on event $\Omega_n^+$:
\begin{eqnarray*}
	\ev_{f_0}\bigg(\|\widehat{g}_{j}-g_{j,0}\|_{L^2}^2|\mathbb{X}\bigg)\leq a_4\sum_{s=1}^d A_{g_{s,0}}^2h_s^{2\beta_s}+\frac{a_4q_+}{n},
	\,\,\,\,\textrm{ for } j=1,2,\ldots, d,
\end{eqnarray*}
and
\begin{eqnarray*}
	\ev_{f_0}(|\widehat{\alpha}-\alpha_0|^2|\mathbb{X})\leq  a_4\sum_{s=1}^d A_{g_{s,0}}^2h_s^{2\beta_s}+\frac{a_4q_+}{n},
\end{eqnarray*}
where $\widehat{\alpha}$ is the estimated coefficient defined in (\ref{eq:additive:alphahat}), and $a_4>0$ is an absolute constant  relying on the density function $Q$ and	 $d$.
\end{lemma}
\begin{proof}[{\it Proof of Lemma \ref{lemma:rate:convergece:individual:additive}}]
Recall $\widehat{f}_{\textrm{pilot}}(\bfx)=\widehat{a}+\sum_{j=1}^d\widehat{f}_j(x_j)=\widehat{\alpha}+\sum_{j=1}^d\widehat{g}_j(x_j)$,
where $\widehat{\alpha}=\widehat{a}+\sum_{j=1}^d\int_0^1 \widehat{f}_j(u)du$ and $\widehat{g}_j(x)=\widehat{f}_j(x)-\int_0^1 \widehat{f}_j(u)du$. By Assumption \ref{A0}
there exists a constant $c>1$ such that for any $g$,
$c^{-1}\int_\Omega g(\bfx)d\bfx\leq \int_\Omega g(\bfx)Q(\bfx)d\bfx\leq c\int_\Omega g(\bfx)d\bfx$. 
By Proposition \ref{proposition:upper:bound:sum:norm} we have
\begin{eqnarray*}
\|\widehat{f}_{\textrm{pilot}}-f_0\|_{L^2}^2&=&\|\widehat{\alpha}-\alpha_0+\sum_{j=1}^d(\widehat{g}_j-g_{j,0})\|_{L^2}^2\\
&\geq&c^{-1}\|\widehat{\alpha}-\alpha_0+\sum_{j=1}^d(\widehat{g}_j-g_{j,0})\|_{{L^2}}^2\\
&\geq&c^{-1}a_3^d\left(|\widehat{\alpha}-\alpha_0|^2+\sum_{j=1}^d\|\widehat{g}_{j}-g_{j,0}\|_{L^2}^2\right)\\
&\geq&c^{-2}a_3^d\left(|\widehat{\alpha}-\alpha_0|^2+\sum_{j=1}^d\|\widehat{g}_{j}-g_{j,0}\|_{L^2}^2\right),
\end{eqnarray*}
where $a_3$ is the constant in Proposition  \ref{proposition:upper:bound:sum:norm}.
By Lemma \ref{lemma:convergence:addtive:sieve} and the above inequality, on event $\Omega_n^+$, the following holds  for any $f_0\in\Lambda_+^{\boldsymbol\beta}(F,\Omega)$:
\begin{eqnarray*}
	\ev_{f_0}\left(\|\widehat{g}_{j}-g_{j,0}\|_{L^2}^2|\mathbb{X}\right)\leq c^2a_3^{-d}\ev_{f_0}\left( \|\widehat{f}_{\textrm{pilot}}-f_0\|_{L^2}^2|\mathbb{X}\right)\leq c^2a_3^{-d} 2^{d+3} \sum_{s=1}^d A_{g_{s,0}}^2h_s^{2\beta_s}+\frac{8c^2a_3^{-d}q_+}{n},
\end{eqnarray*}
for $j=1,2,\ldots, d$,
and
\begin{eqnarray*}
	\ev(|\widehat{\alpha}-\alpha_0|^2|\mathbb{X})\leq c^2a_3^{-d} 2^{d+3} \sum_{s=1}^d A_{g_{s,0}}^2h_s^{2\beta_s}+\frac{8c^2a_3^{-d}q_+}{n}.
\end{eqnarray*}
Therefore, the desired results follow with $a_4=c^2a_3^{-d} 2^{d+3}$.
Proof is complete.
\end{proof}

Given previous Lemmas, we are ready to prove Theorem \ref{thm:rate:convergence:additive:main:text}. By Lemma \ref{lemma:spline:approximation}, it holds that 
$$\sup_{x \in [0, 1]}|C_j^T\bfB_{k_j,j}(x)-g_{j,0}(x)|\leq A_{g_{j,0}}h_j^{\beta_j}$$ for some $C_j\in \mathbb{R}^{M_j+k_j-1}$ with $\|C_{j}\|_\infty\leq A_{g_{j,0}}$. Let $g_{j}^*=C_j^T\bfB_{k_j,j}$
for $j=1,\ldots,d$. Recall that  $\widehat{g}_j$ can be written as $\widehat{C}_j^T\bfB_{k_j,j}(x)$ for some $\widehat{C}_j\in \mathbb{R}^{M_j+k_j-1}$ and the neural network approximating the additive component is $\widetilde{g}_j(x)=\widehat{C}_j^T\widetilde{\bfB}_{k_j,j}(x)$ according to (\ref{eq:expression:fnet:additive}).

By Lemma \ref{lemma:rate:convergece:individual:additive}, for any $f_0\in\Lambda_+^{\boldsymbol\beta}(F,\Omega)$ we have
	\begin{eqnarray}
		\ev_{f_0}\left(\|\widehat{g}_{j}-g_{j,0}\|_{L^2}^2|\mathbb{X}\right)&\leq& a_4\sum_{s=1}^dA_{g_{s,0}}^2 h_s^{2\beta_s}+\frac{a_4q_+}{n}, \textrm{ for } j=1,2,\ldots, d,\nonumber\\
		\;\ev_{f_0}\left(|\widehat{\alpha}-\alpha_0|^2|\mathbb{X}\right)&\leq& a_4\sum_{s=1}^dA_{g_{s,0}}^2 h_s^{2\beta_s}+\frac{a_4q_+}{n},\label{eq:thm:rate:convergence:additive:1}
	\end{eqnarray}
where  $a_4$ is the constant in Lemma  \ref{lemma:rate:convergece:individual:additive}.
By Lemma \ref{lemma:eigen:value:design:matrix}, for every $j=1,\ldots,d$ we have
\begin{eqnarray*}
a_1h_j(\widehat{C}_j-C_j)^T(\widehat{C}_j-C_j)&\leq&\int |\widehat{C}_j^T\bfB_{k_j,j}(x_j)-C_j^T\bfB_{k_j,j}(x_j)|^2Q(\bfx)d\bfx\\
&=&\|\widehat{g}_j-g^*_j\|^2\\
&\leq&2\|\widehat{g}_j-g_{j,0}\|_{L^2}^2+2\|{g}_j^*-g_{j,0}\|_{L^2}^2,
\end{eqnarray*}
which further implies that the following holds on $\Omega_n^+$:
\begin{eqnarray*}
	\ev_{f_0}\left(\widehat{C}_j^T\widehat{C}_j|\mathbb{X}\right)&\leq& 2C_j^TC_j+2\ev_{f_0}\left((\widehat{C}_j-C_j)^T(\widehat{C}_j-C_j)|\mathbb{X}\right)\\
	&\leq& 2q_+A_{g_{j,0}}^2+4a_1^{-1}h_j^{-1}\ev_{f_0}\left(\|\widehat{g}_j-g_{j,0}\|_{L^2}^2|\mathbb{X}\right)+ 4a_1^{-1}h_j^{-1}\ev_{f_0}\left(\|{g}_j^*-g_{j,0}\|_{L^2}^2|\mathbb{X}\right)\\
	&\leq&2q_+A_{g_{j,0}}^2+   4a_1^{-1}h_j^{-1} \bigg( a_4\sum_{s=1}^dA_{g_{s,0}}^2 h_s^{2\beta_s}+\frac{a_4q_+}{n}\bigg)+ 4a_1^{-1}h_j^{-1}A_{g_{j,0}}^2h_j^{2\beta_j}\\
	&\leq& (2A_{g_{j,0}}^2+4a_1^{-1}a_4)\bigg(q_++\frac{q_+}{nh_j}\bigg)+4a_1^{-1} (a_4+1) h_j^{-1}\sum_{s=1}^dA_{g_{s,0}}^2 h_s^{2\beta_s}\\
	&\leq& a_6 \left(\sum_{v=1}^d A_{g_{v,0}}^2+a_1^{-1}a_4\right)\left(q_++h_j^{-1}\sum_{s=1}^dh_s^{2\beta_s}\right),
\end{eqnarray*}
with $a_6=8+8a_1^{-1} (a_4+1) $. In the last inequality we have used $nh_j\to \infty$.
Recall $\widetilde{g}_j=\widehat{C}_j^T\widetilde{\bfB}_{k_j,j}(x)$. Therefore, Lemma \ref{lemma:approximation:b:spline:d:1} implies that the following holds on event $\Omega_n^+$: $\frac{8^{k}}{14}4^{-m}$
\begin{eqnarray*}
	\ev_{f_0}(\|\widetilde{g}_j-\widehat{g}_j\|_{L^2}^2|\mathbb{X})&=&\ev_{f_0}\left(\|\widehat{C}_j^T\widetilde{\bfB}_{k_j,j}-\widehat{C}_j^T{\bfB}_{k_j,j}\|_{L^2}^2|\mathbb{X}\right)\\
	&\leq& (M_j+k_j-1)\ev_{f_0}\left(\widehat{C}_j^T\widehat{C_j}|\mathbb{X}\right)\sup_{x\in [0,1]}
	\|\widetilde{\bfB}_{k_j,j}(x)-{\bfB}_{k_j,j}(x)\|_\infty^2\\
	&\leq&a_6 \left(\sum_{v=1}^d A_{g_{v,0}}^2+a_1^{-1}a_4\right) 64^{k_j+1}(M_j+k_j-1)\left(q_++h_j^{-1}\sum_{s=1}^dh_s^{2\beta_s}\right)16^{-m}.
\end{eqnarray*}
By the above inequality and (\ref{eq:thm:rate:convergence:additive:1}), on event $\Omega_n^+$, we have
\begin{eqnarray*}
	&&\ev_{f_0}(\|\widetilde{g}_j-g_{j,0}\|_{L^2}^2|\mathbb{X})\\&\leq &2	\ev_{f_0}(\|\widetilde{g}_j-\widehat{g}_j\|_{L^2}^2|\mathbb{X})+	2\ev_{f_0}(\|\widehat{g}_j-g_{j,0}\|_{L^2}^2|\mathbb{X})\\
	&\leq&2a_6 \left(\sum_{v=1}^d A_{g_{v,0}}^2+a_1^{-1}a_4\right) 64^{k_j+1}(M_j+k_j-1)\left(q_++h_j^{-1}\sum_{s=1}^dh_s^{2\beta_s}\right)16^{-m}\\
	&&+2a_4\sum_{s=1}^dA_{g_{s,0}}^2 h_s^{2\beta_s}+\frac{a_4q_+}{n}
\end{eqnarray*}
As a consequence, on event $\Omega_n^+$, it follows that
\begin{eqnarray*}
	&&\ev_{f_0}(\|\widehat{f}_{\textrm{net}}-f_{0}\|_{L^2}^2|\mathbb{X})\\
	&\leq& 2^d\ev(|\widehat{\alpha}-\alpha_0|^2|\mathbb{X})+2^d\sum_{j=1}^d\ev(\|\widetilde{g}_j-g_{j,0}\|_{L^2}^2|\mathbb{X})\\
	&\leq&	2^da_4 \bigg(\sum_{s=1}^dA_{g_{s,0}}^2 h_s^{2\beta_s}+\frac{q_+}{n}\bigg)\\
	&&+2^{d+1}a_6  \sum_{j=1}^d\left(\sum_{v=1}^d A_{g_{v,0}}^2+a_1^{-1}a_4\right) 64^{k_j+1}(M_j+k_j-1)\left(q_++h_j^{-1}\sum_{s=1}^dh_s^{2\beta_s}\right)16^{-m}\\
	&&+2^{d+1}a_4d\sum_{s=1}^dA_{g_{s,0}}^2 h_s^{2\beta_s}+\frac{a_4dq_+}{n}.
\end{eqnarray*}
Since $q_+=\sum_{j=1}^d (M_j+k_j-1)\asymp \sum_{j=1}^d M_j$, $h_j\asymp M_j^{-1}$ and  $\sup_{g_{j,0}\in \Lambda^{\beta_j}(F, [0, 1])}A_{g_{j,0}}<\infty$ by Lemma \ref{lemma:spline:approximation}, 
taking supremum of the above inequality leads to
\begin{eqnarray*}
\sup_{f_0\in \Lambda^{\bm{\beta}}_+(F, \Omega)}\ev_{f_0}(\|\widehat{f}_{\textrm{net}}-f_{0}\|_{L^2}^2|\mathbb{X})=O_P\bigg(\sum_{j=1}^d M_j^{-2\beta_j}\bigg)+O_P\bigg(\sum_{j=1}^d\frac{M_j}{n}\bigg)+O_P\bigg(\sum_{j=1}^d M_j^2 4^{-2m}\bigg).
\end{eqnarray*}
Using (\ref{eq:counting:fnet:additive}), we know  $L=(2m+3)\max_{1\leq j\leq d}(k_j+1)+1$ and $T=\sum_{j=1}^d2^{k_j+4}(M_j+2k_j)$. The above inequality further leads to
\begin{eqnarray*}
\inf_{\widehat{f}\in\mathcal{F}(L,\bfp(T))}\sup_{f_0\in \Lambda^{\bm{\beta}}_+(F, \Omega)}\ev_{f_0}(\|\widehat{f}_{\textrm{net}}-f_{0}\|_{L^2}^2|\mathbb{X})
&\leq&   \sup_{f_0\in \Lambda^{\bm{\beta}}_+(F, \Omega)}\ev_{f_0}(\|\widehat{f}_{\textrm{net}}-f_{0}\|_{L^2}^2|\mathbb{X})\\
&=&O_P\bigg(T^{-2\beta_*}+\frac{T}{n}+T^2 4^{-2m}\bigg).
\end{eqnarray*}
We can always choose $k_j=\floor{\beta}_j+1$ for $j=1,\ldots, d$. Therefore the integer $k\geq \max(\beta_1,\ldots, \beta_d, 2)$ implies $k\geq \max(k_1,\ldots, k_d, 2)$ and   $L=(2m+3)\max_{1\leq j\leq d}(k_j+1)+1\leq 2m(k+1)+3(k+1)+1$. Substituting $m$ with $L$, we complete the proof.

\textbf{Acknowledgement.} The authors would like to thank the Editor and an anonymous reviewer for their constructive suggestions that have led to a significant improvement in the manuscript. Zuofeng Shang acknowledges supports by NSF DMS-1764280 and DMS-1821157.

%
\bibliography{GFERef}{}
\bibliographystyle{apalike}

\clearpage
\setcounter{subsection}{0}
\renewcommand{\thesubsection}{A.\arabic{subsection}}
\setcounter{subsubsection}{0}
\renewcommand{\thesubsubsection}{\textbf{A.\arabic{subsection}.\arabic{subsubsection}}}
\setcounter{equation}{0}
\renewcommand{\theequation}{A.\arabic{equation}}
\setcounter{lemma}{0}
\renewcommand{\thelemma}{A.\arabic{lemma}}
\setcounter{proposition}{0}
\renewcommand{\theproposition}{A.\arabic{proposition}}
\setcounter{page}{1}
\section*{Appendix}
\subsection{Proof of Lemma \ref{lemma:spline:approximation}}
In this subsection, we provide the proof of Lemma \ref{lemma:spline:approximation}. For simplicity, we consider the case with $d=2$. The extension to the scenario with $d>2$ can be done similarly.

Given integers $k, M\geq 2$ and knots $t_{-k+1}<t_{-k+2}<\ldots< t_0<t_1< \ldots<t_M< t_{M+1} <\ldots< t_{M+k-1}$ with 
$t_0=0, t_M=1$. Since $d=2$, we can relabel the tensor product B-spline basis as $B_{i,k}(x_1)B_{j,k}(x_2)$, for $(x_1, x_2)^T\in \Omega$ and $i,j=-k+1,\ldots, M-1$. We would like to comment that the basis is denoted as $D_{\bfi,k}$ in previous section. As a consequence, the function space spanned by $B_{i,k}(x_1)B_{j,k}(x_2)$ is defined as
$$\Theta_n=\{f(\bfx)=\sum_{i=-k+1}^{M-1}\sum_{j=-k+1}^{M-1}c_{ij}B_{i,k}(x_1)B_{j,k}(x_2) | c_{ij}\in \mathbb{R} \textrm{ and } \bfx=(x_1, x_2)^T\in \Omega\}.$$

Let us borrow some definition from the Section 15.1 in \cite{gkkw06}. Let $\mcC$ be the collection of continuous function supported on $\Omega$. A linear operator  $\kappa: \mcC\to \Theta_n$ is called a quasi interpolant if
\begin{eqnarray*}
\kappa f(\bfx)=\sum_{i=-k+1}^{M-1}\sum_{j=-k+1}^{M-1}\kappa_{ij}(f) B_{i,k}(x_1)B_{j,k}(x_2),
\end{eqnarray*}
where $\kappa_{ij}(f)$ is a constant depending only on the values of $f$ in $[t_i, t_{i+k})\times [t_j, t_{j+k})$. Moreover, $\kappa$ is said to have order $k$ if  $\kappa f=f$ for all polynomial $f$ with the degrees of $x_1$ and $x_2$ not greater than $k-1$.
\begin{lemma}\label{lemma:bound:kappa}[Theorem 15.2 of \cite{gkkw06}]
Given integers $k, M\geq 2$ and knots $t_{-k+1}<t_{-k+2}<\ldots< t_0<t_1< \ldots<t_M< t_{M+1} <\ldots< t_{M+k-1}$ with 
$t_0=0, t_M=1$.  There exists a quasi interpolant  $\kappa: \mcC\to \Theta_n$ with order $k$ such that
\begin{eqnarray*}
|\kappa_{ij}(f)|\leq L_k \sup_{\bfx \in [t_i, t_{i+k})\times [t_j, t_{j+k})} |f(\bfx)|.
\end{eqnarray*}
Here $L_k$ is a constant depending only on $k$ but not on the knots.
\end{lemma}

We are ready to prove Lemma  \ref{lemma:spline:approximation}.  Suppose $f\in \Lambda^{\beta}(F, \Omega)$. For fixed $\bfu\in [t_i, t_{i+k})\times [t_j, t_{j+k})$, let us define the following local Taylor polynomial:
\begin{eqnarray*}
p_\bfu(\bfx)=\sum_{|\bm{\alpha}|\leq \floor{\beta}} \partial^{\bm{\alpha}}f(\bfu)\frac{(\bfx-\bfu)^{\bm{\alpha}}}{\bm{\alpha}!}\quad \textrm{ for } \bfx \in [t_i, t_{i+k})\times [t_j, t_{j+k}).
\end{eqnarray*}
By Taylor's theorem, it follows that
\begin{eqnarray*}
f(\bfx)=\sum_{|\bm{\alpha}|< \floor{\beta}} \partial^{\bm{\alpha}}f(\bfu)\frac{(\bfx-\bfu)^{\bm{\alpha}}}{\bm{\alpha}!}+\sum_{|\bm{\alpha}|=\floor{\beta}}\frac{\floor{\beta}}{\bm{\alpha}!}(\bfx-\bfu)^{\bm{\alpha}}\int_0^1 (1-t)^{\floor{\beta}-1}\partial^{\bm{\alpha}}f(\bfu+t(\bfx-\bfu))dt.
\end{eqnarray*}
Suppose $\bfu=(u_1,u_2)^T, \bfx=(x_1, x_2)^T \in [t_i, t_{i+k})\times [t_j, t_{j+k})$, then Assumption \ref{Assumption:A1} implies that $\|\bfu-\bfx\|\leq \sqrt{2(kh)^2}\leq 2kh$. Let us consider two cases of $\beta$.

Case 1: If $\floor{\beta}=0$, then $p_\bfu(\bfx)=f(\bfu)$. By the definition of $\Lambda^{\beta}(F, \Omega)$, it follows that
\begin{eqnarray*}
|f(\bfx)-p_\bfu(\bfx)|=|f(\bfx)-f(\bfu)|\leq F\|\bfx-\bfu\|^{\beta}\leq F(2k)^{\beta}h^{\beta}.
\end{eqnarray*}

Case 2: If $\floor{\beta}\geq 1$, then $\int_0^1 (1-t)^{\floor{\beta}-1}dt=1/{\floor{\beta}}$. Therefore, we have
\begin{eqnarray*}
|f(\bfx)-p_\bfu(\bfx)|&\leq &\sum_{|\bm{\alpha}|=\floor{\beta}}\bigg|\frac{\floor{\beta}}{\bm{\alpha}!}(\bfx-\bfu)^{\bm{\alpha}}\bigg| \int_0^1 (1-t)^{\floor{\beta}-1}\bigg|\partial^{\bm{\alpha}}f(\bfu+t(\bfx-\bfu))-\partial^{\bm{\alpha}}f(\bfu)\bigg|dt\\
&\leq& \sum_{|\bm{\alpha}|=\floor{\beta}} \floor{\beta}  |x_1-u_1|^{\alpha_1}|x_2-u_2|^{\alpha_2} \int_0^1 (1-t)^{\floor{\beta}-1} F\|t(\bfx-\bfu)\|^{\beta-\floor{\beta}}dt\\
&\leq&F \floor{\beta} \sum_{|\bm{\alpha}|=\floor{\beta}} (kh)^{\alpha_1}(kh)^{\alpha_2} \|\bfx-\bfu\|^{\beta-\floor{\beta}}\int_0^1 (1-t)^{\floor{\beta}-1}dt\\
&\leq& F\sum_{|\bm{\alpha}|=\floor{\beta}}k^{\floor{\beta}} h^{\floor{\beta}} (2kh)^{\beta-\floor{\beta}}\\
&\leq& F(\floor{\beta}+1)^2(2k)^{\beta}h^{\beta}.
\end{eqnarray*}

Combining the above two cases, we show that
\begin{eqnarray*}
|f(\bfx)-p_\bfu(\bfx)|\leq  F(\floor{\beta}+1)^2(2k)^{\beta}h^{\beta},
\end{eqnarray*}
for all $\bfx, \bfu \in [t_i, t_{i+k})\times [t_j, t_{j+k})$ and $f\in \Lambda^{\beta}(F, \Omega)$. Since the operator $\kappa$ is linear, and $p_u(\bfx)$ is a polynomial with degrees of $x_1$ and $x_2$ not greater than $\floor{\beta}$. Since $\kappa$ is an interpolant with order $k$ by Lemma  \ref{lemma:bound:kappa}, and $p_\bfu$ is a polynomial with degree at most $\floor{\beta}$, the condition $k\geq \beta$ implies that $k-1\geq \floor{\beta}$. As a consequence, it follows that
\begin{eqnarray*}
|\kappa [f(\bfx)- p_\bfu(\bfx)]|&=&|\sum_{i=-k+1}^{M-1}\sum_{j=-k+1}^{M-1}\kappa_{ij}(f-p_\bfu) B_{i,k}(x_1)B_{j,k}(x_2)|\\
&\leq& \sum_{i=-k+1}^{M-1}\sum_{j=-k+1}^{M-1}|\kappa_{ij}(f-p_\bfu)| B_{i,k}(x_1)B_{j,k}(x_2)\\
&\leq& \sup_{-k+1\leq i\leq M-1} \sup_{-k+1\leq j\leq M-1} |\kappa_{ij}(f-p_\bfu)| \\
&\leq&  \sup_{-k+1\leq i\leq M-1} \sup_{-k+1\leq j\leq M-1} L_k \sup_{\bfv\in [t_i, t_{i+k})\times [t_j, t_{j+k})}|f(\bfv)-p_\bfu(\bfv)|\\
&\leq& L_k F(\floor{\beta}+1)^2(2k)^{\beta}h^{\beta} \quad \textrm{ for all } \bfx \in \Omega.
\end{eqnarray*}
Combining the above inequality, we conclude that
\begin{eqnarray*}
|\kappa f(\bfx)-f(\bfx)|&\leq &|\kappa f(\bfx)-p_\bfu(\bfx)|+|p_\bfu(\bfx)-f(\bfx)|\\
&=&|\kappa f(\bfx)-\kappa p_\bfu(\bfx)|+|p_\bfu(\bfx)-f(\bfx)|\\
&=&|\kappa [f(\bfx)- p_\bfu(\bfx)]|+|p_\bfu(\bfx)-f(\bfx)|\\
&\leq& (L_k+1)F(\floor{\beta}+1)^2(2k)^{\beta}h^{\beta} \quad \textrm{ for all } \bfx \in \Omega.
\end{eqnarray*}
Notice that $\kappa f(\bfx)=\sum_{i=-k+1}^{M-1}\sum_{j=-k+1}^{M-1}\kappa_{ij}(f) B_{i,k}(x_1)B_{j,k}(x_2)=\sum_{\bfi \in \Gamma}c_\bfi D_{\bfi, k}(\bfx)$, where $c_\bfi$'s is the sequence  $\kappa_{ij}(f)$'s after relabelling. Using Lemma \ref{lemma:bound:kappa} again, we show that $|c_{\bfi}|\leq L_k\|f(\bfx)\|_{\sup}$.  Clearly, we can choose 
\begin{eqnarray*}
A_f=(L_k+1)F(\floor{\beta}+1)^2(2k)^{\beta}+L_k\|f(\bfx)\|_{\sup},
\end{eqnarray*}
which satisfies $\sup_{f\in  \Lambda^{\beta}(F, \Omega)}A_f\leq (L_k+1)F(\floor{\beta}+1)^2(2k)^{\beta}+L_kF<\infty$. The proof is complete.

\subsection{Index of Symbols}
\begin{itemize}
\item $k$:  the smallest integer satisfying $k\geq \max(\beta, 2)$ for Theorems 1-3 and $k\geq \max(\beta_1,\ldots, \beta_d, 2)$ for Theorem 4. 
\item $m$: a diverging auxiliary variable for the number of hidden layers, which is related to the construction of network product operator; see Lemma \ref{proposition:appriximation:product}, (\ref{eq:counting:fnet}) and (\ref{eq:counting:fnet:additive}).     
\item  $M$: a diverging auxiliary variable for the number of nodes in each hidden layer, which is related to the number of knots for B-spline basis; see Section \ref{section:pilot:bspline} and  (\ref{eq:counting:fnet}).      
\item $h$: $h=M^{-1}$,  knots separation distance; see Assumption \ref{Assumption:A1}.
\item $q$: $q=(M+k-1)^d$, number of  tensor product B-spline basis functions; see Section \ref{section:pilot:bspline}.
\item $\Theta_n$:  a  function space spanned by   tensor product B-spline basis; see Section \ref{sec:aymptpt:pilot}.
\item $\Omega_n$: an event with probability approaching one; see  (\ref{eq:definition:Omegan}).
\item $a_1, a_2$:    universal constants relying on $k$ and the density $Q$; see Lemma \ref{lemma:eigen:value:design:matrix}. 
\item $a_3$: a universal constant relying on the density  $Q$; see Proposition \ref{proposition:upper:bound:sum:norm}.           
\item $a_4$:  a universal constant  relying on $d$ and the density     $Q$; see Lemma \ref{lemma:rate:convergece:individual:additive}.
\item $k_j$:      a fixed constant indicating the order of B-spline basis for additive model, which requires $k_j\geq \beta_j$; see Assumption \ref{Assumption:A2}.
\item $M_j$:    a diverging auxiliary variable for the number of nodes in each hidden layer for additive model, which is related to the number of knots for B-spline basis; see Assumption \ref{Assumption:A2} and (\ref{eq:counting:fnet:additive}).  
\item $h_j$:  $h_j=M_j^{-1}$,  knots separation distance    for additive model; see Assumption \ref{Assumption:A2}.
\item $q_+$:  $q=1+\sum_{j=1}^d(M_j+k_j-2)$, number of   B-spline basis functions for additive model; see (\ref{eq:symbol:additive}).
\item $\Theta_n^+$: function space spanned by B-spline basis for additive model; see (\ref{eq:symbol:additive}).
\item $\Omega_n^+$: an event with probability approaching one; see  (\ref{eq:symbol:additive}).
\end{itemize}
\end{document}